\pdfoutput=1
\documentclass{article}

\PassOptionsToPackage{numbers, compress}{natbib}

\usepackage[preprint]{neurips_2025}

\usepackage[utf8]{inputenc} 
\usepackage[T1]{fontenc}    
\usepackage{hyperref}       
\usepackage{url}            
\usepackage{booktabs}       
\usepackage{amsfonts}       
\usepackage{nicefrac}       
\usepackage{microtype}      
\usepackage{xcolor}         
\usepackage{graphicx}
\usepackage{subcaption}
\usepackage{subfloat}
\usepackage[inline]{enumitem}
\usepackage{amsmath}
\usepackage{listing}
\usepackage{listings}
\usepackage[breakable]{tcolorbox}
\usepackage{changepage}
\usepackage{wrapfig}

\title{GIFARC: Synthetic Dataset for Leveraging Human-Intuitive Analogies to Elevate AI Reasoning}

\author{
  Woochang Sim \\
  GIST \\
  \small{\texttt{woochang@gm.gist.ac.kr}} \\
  \And
  Hyunseok Ryu \\
  GIST \\
  \small{\texttt{omnyx2@gmail.com}} \\
  \And
  Kyungmin Choi \\
  GIST \\
  \small{\texttt{youvegotmail@gm.gist.ac.kr}} \\
  \And
  Sungwon Han \\
  GIST \\
  \small{\texttt{lion4152@gmail.com}} \\
  \And
  Sundong Kim \\
  GIST \\
  \small{\texttt{sundong@gist.ac.kr}} \\
  }

\begin{document}

\maketitle

\begin{abstract}
\label{Sec:Abstract}
The Abstraction and Reasoning Corpus (ARC) poses a stringent test of general AI capabilities, requiring solvers to infer abstract patterns from only a handful of examples. 
Despite substantial progress in deep learning, state-of-the-art models still achieve accuracy rates of merely 40–55\% on 2024 ARC Competition, indicative of a significant gap between their performance and human-level reasoning. 
In this work, we seek to bridge that gap by introducing an analogy-inspired ARC dataset, GIFARC\footnote{Full dataset is available on Hugging Face at ~\url{https://huggingface.co/datasets/DumDev/gif_arc}. \newline All 10,000 tasks generated with GIFARC are visualized in \url{https://gifarc.vercel.app/}.}.
Leveraging large language models (LLMs) and vision-language models (VLMs), we synthesize new ARC-style tasks from a variety of GIF images that include analogies. 
Each new task is paired with ground-truth analogy, providing an explicit mapping between visual transformations and everyday concepts. 
By embedding robust human-intuitive analogies into ARC-style tasks, GIFARC guides AI agents to evaluate the task analogically before engaging in brute-force pattern search, thus efficiently reducing problem complexity and build a more concise and human-understandable solution.
We empirically validate that guiding LLM with analogic approach with GIFARC affects task-solving approaches of LLMs to align with analogic approach of human. \looseness=-1 

\end{abstract}
\section{Introduction}
\label{Sec:Intro}

Recent advances in AI have produced impressive results in specialized domains, ranging from image classification to natural language processing. 
Despite these achievements, many of these systems remain limited in their capacity for genuinely flexible, general-purpose reasoning -- a core characteristic often associated with artificial general intelligence (AGI). 
In this context, the Abstraction and Reasoning Corpus (ARC) was introduced as a benchmark aiming to measure the ability of artificial general intelligence systems to abstract visual patterns and reason about them in a manner akin to human cognition~\cite{chollet2019ARC}. 
ARC tasks are intentionally designed to challenge AI the ability of high-level abstract reasoning by requiring solutions derived from minimal few-shot examples. 
This requirement closely mirrors the cognitive flexibility inherent to human task-solving, thus making ARC an essential dataset for probing the fundamental limitations and capabilities of current AI systems in relation to achieving human-level AGI.

Despite the excitement surrounding deep learning, current best-performing models in the 2024 ARC Prize competition left a significant gap compared to human-level proficiency~\cite{lab422024arcprize}. 
While traditional deep neural networks typically excel with ample data, they struggle in example-deficient tasks that require understanding. This shortcoming underscores the limitations of purely data-driven approaches, especially when the underlying task involves rich forms of abstract and compositional reasoning~\cite{dziri2023faith, lee2024reasoning}. As a result, ARC provides strong evidence that deep learning alone does not yet capture the breadth of human reasoning, pointing to critical shortcomings in its capacity for generalization, pattern inference, and systematic task-solving.

\begin{figure}[t!]
    \centering
    \includegraphics[width=\columnwidth]{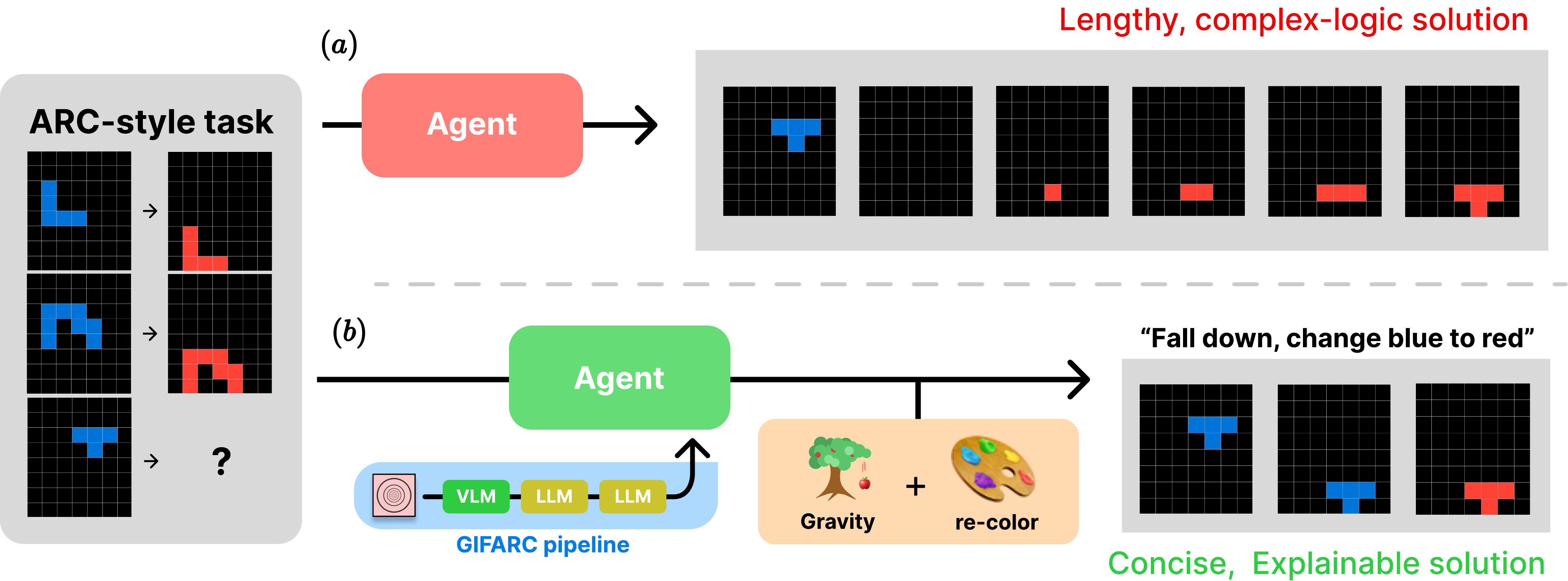}
    \caption{Illustration of how differently an agent solves ARC-style task when it is guided with or without analogic approach. In (a), an agent attempts to solve the ARC-style task in brute-force manner and is in risk to result logically complex solutions. In contrast, When the agent has been guided to think analogically with GIFARC dataset, as depicted in (b), it finds out a more concise and human-intuitive solution.}
    \label{fig:desired_enhancement}
\end{figure}

Why does this gap exist? One crucial factor lies in how humans leverage analogies to map new visual or conceptual clues onto existing knowledge.
When people see a grid transformation as in Figure~\ref{fig:desired_enhancement}, they can instantly map this visual pattern onto familiar concepts such as gravity and color change.
These priors, accumulated from a lifetime of diverse visual and conceptual experiences, provide immediate insight into the task constraints and sharply reduce the complexity of the tasks. 
Rather than searching blindly through every possible transformation, humans can reject lengthy or counterintuitive candidates and focus on the small subset of patterns consistent with their analogies.

In contrast, deep learning models rarely have an explicit mechanism for analogy-making~\cite{mitchell2021abstraction}. 
Even advanced neural networks typically rely on statistical correlations learned from large training sets, lacking the means to ``tag'' or ``reframe'' new tasks in terms of previously understood scenarios~\cite{bober2024neural}. 
Consequently, these models miss a powerful heuristic: an analogy-driven search strategy that narrows down the solution space by linking visual abstractions to real-world events. 

\paragraph{Need of analogy-contained dataset.} 
To help bridge this gap between human cognition and AI, the agent should obtain human-level amount of analogies in prior to any task-specific training. 
Since the models we aim to train are predominantly data-driven, the need for training dataset that consists of useful analogies naturally arise. 
The dataset we seek contains following criteria; 

\begin{enumerate}[leftmargin=*,label=(\roman*)]
\item ~\emph{ARC-style examples} : the dataset should contain analogies in forms of ARC-style examples. 
Since the agent is evaluated via ARC benchmark, the agent has the world perception in form of 2D grids. 
In addition, the analogies one needs to know should be expressible onto an ARC-style 2D grid. 
Thus, the analogy we intend to train should also be expressed and given in ARC-style 2D grid in order for the agents to understand and utilize during evaluation. 

\item ~\emph{Executable codes} : In order for the agent to capture how to manipulate 2D grid in accordance to the analogy, analogy should also be expressed in executable codes. 
Therefore, the analogy of transformation should also be given as lines of executable codes. 
With those 2D grids and codes, the agent should be able to correlate ARC-style transformation with its underlying analogy and find it during evaluation. 

\end{enumerate}

In this paper, we propose an analogy-inspired synthetic ARC dataset, GIFARC. 
Drawing on large language model (LLM) and vision-language model (VLM), we developed a framework to synthesize ARC-style tasks from online GIF images by extracting various visual patterns and the analogies they evoke. 
These GIFs encompass a range of scenarios (e.g., snowfall, blooming flowers, moving objects), each providing a different type of visual transformation. 
We encode these transformations in newly generated ARC-style tasks and, crucially, supply ground-truth analogy labels that clarify the underlying conceptual mapping (e.g., “water flow blocked by an obstacle”). 

\paragraph{Role of GIF as the basis of ARC-style task generation.} 
Using GIFs as the starting ingredient of ARC-style tasks, our GIFARC synthesis pipeline takes great advantages in picking human-understandable and visualizable analogies, as GIFs are series of images that are intentionally drawn to visualize certain ideas. In addition, GIF also has advantages in holding a sufficiently small number of analogies. In terms of analogy extraction simplicity, it is critical for the analogies to be implied in the image series without any entanglement to one another. Longer image series such as videos are more likely to hold excessive number of analogies and consequently have a higher risk of entanglement in between. Thus, GIF is an adequate form of analogy-implying image series.

These new ARC-style tasks generated from GIFs will possess a variety of intuitive analogies. We expect that the agent trained with this dataset can gain the grasp of analogical approaches and utilize them in the solving step. 
By incorporating diverse analogies into the dataset, we hope to facilitate models that can “think analogically” before searching for solutions, thereby reducing the complexity of task-solving.
In addition, by converting the visual data to verbose expression before generating the ARC-style tasks, we expect that the agent will specifically gain the ability to verbally point out the analogy intended in the ARC-style task. \looseness=-1 

Our GIFARC dataset synthesis pipeline operates in three sequential stages.
First, a VLM converts raw GIF into a structured JSON record listing the objects in the scene, invariant backgrounds, dynamic trajectories, and concise analogy.
LLM then compresses that verbose abstraction into a task sketch comprising concepts and descriptions for ARC-style tasks.
Conditioned on the sketch, LLM finally generate executable ARC-style tasks via retrieval-based in-context learning.

Running this pipeline on motion-rich GIFs produces 10k unique high-quality ARC-style tasks spanning 20 analogy categories. 
We empirically demonstrated that the analogy information in our GIFARC dataset guides LLMs to reason through ARC tasks in a manner more aligned with human-like reasoning.

\section{Related Work}
\paragraph{Abstraction and Reasoning Corpus.}
The Abstraction and Reasoning Corpus (ARC), introduced by Chollet (2019)~\cite{chollet2019ARC}, is an open-ended benchmark designed to evaluate broad generalization from minimal demonstrations. 
Its goal is to test how efficiently an agent can acquire and apply abstract patterns with just a few examples. ARC-AGI-1 (often referred to as ``ARC-1'') quickly became a focal point in the research community. 
The initial Kaggle competition in 2020 yielded a top solve rate of only 21\%, significantly below human-level performance. 
With recent advances in large language models, OpenAI o3-preview model scored 75\% on ARC-AGI-1 using low level of computation and reached 87\% accuracy with higher level of computation, marking the first effective solution to the challenge~\cite{pfister2025understanding}. 
This milestone sparked interest and led to the development of ARC-AGI-2~\cite{arc2025}, which was designed to pose even greater challenges while preserving accessibility to humans. 

\paragraph{Synthetic ARC Generation.}
To address the data scarcity inherent in ARC, which disturbs both symbolic program synthesis and the training of neural models, researchers have turned to generating programmatic data using human-curated, AI-generated, or hybrid pipelines~\cite{assouel2022object,legris2024h,opielka2024large,qi2021pqa}.
One such effort is ReARC, which manually scripts generative procedures for each task to produce an effectively infinite set of in-distribution examples~\cite{hodel2024rearc}.
LARC (Language-Complete ARC) crowd-sources natural language explanations of tasks, enabling models to learn from natural programs instead of raw grids~\cite{acquaviva2022communicating}.
Other benchmarks such as ConceptARC, Mini-ARC, and 1D-ARC use manually crafted rules to produce diverse yet interpretable task distributions~\cite{kim2022playgrounds,moskvichev2023conceptARC,xu2024llms}.
Recent approaches automate this process at scale. For instance, BARC generates 400,000 training examples by prompting GPT-4 to mutate a curated set of 160 Python-based solvers~\cite{li2024combining}. 
Each seed includes a transformation function, an input generator, and natural language description.
Similarly, MC-LARC transforms ARC tasks into multiple-choice formats, enabling scalable evaluation of language models' intermediate reasoning stages~\cite{shin2024mclarc}. \looseness=-1 

Different from existing in-distribution resampling or code-mutation pipelines, our work introduces GIFARC, a new dataset that mines human-intuitive GIFs, distills their analogical structure with vision-language models, and auto-compiles each analogy into an executable ARC-domain specific language (ARC-DSL) code, explicitly teaching models how to think by analogy rather than simply giving them more examples to memorize.
\section{Method}
\label{sec:method}
\subsection{Overview}
\label{subsec:overview}

\paragraph{Problem formulation.}
Given a set of GIF images $\mathcal{G} = \{ g_j \}_{j=1}^N$, our goal is to construct an analogy-grounded ARC dataset, GIFARC, containing ARC-style tasks and represented as triples
\(
\mathcal{T}\!=\!(\mathcal{E},\alpha,\phi)
\)
where
\begin{enumerate*}[label=(\roman*)]
\item \(\mathcal{E}=\{(x_i,y_i)\}_{i=1}^{K}\) is a set of \(K\) input–output grid pairs of ARC-style task. Here, \(x_i\in\{0,\dots,9\}^{h_{x,i}\times w_{x,i}}\) is the input grid, \(y_i\in\{0,\dots,9\}^{h_{y,i}\times w_{y,i}}\) is the output grid, with \(h_{G,i},w_{G,i}\) respectively being the height and width of the \(G\) grid (either grid \(x\) or \(y\)) in \(i\)-th pair of \(\mathcal{E}\);
\item \(\alpha\) is a short natural-language analogy description (e.g.,\ “blocked water flow”).
\item \(\phi\) is a Python program implementing \(\mathcal{F}:\mathcal{X}\!\to\!\mathcal{Y}\), the latent deterministic transformation such that \(y_i=\mathcal{F}(x_i)\);
\end{enumerate*}
GIFARC $(\mathcal{D}=\{\mathcal{T}_j\}_{j=1}^{N})$ therefore pairs each ARC task with its provenance and analogy.

\begin{figure}[t!]
    \centering
    \includegraphics[width=\columnwidth]{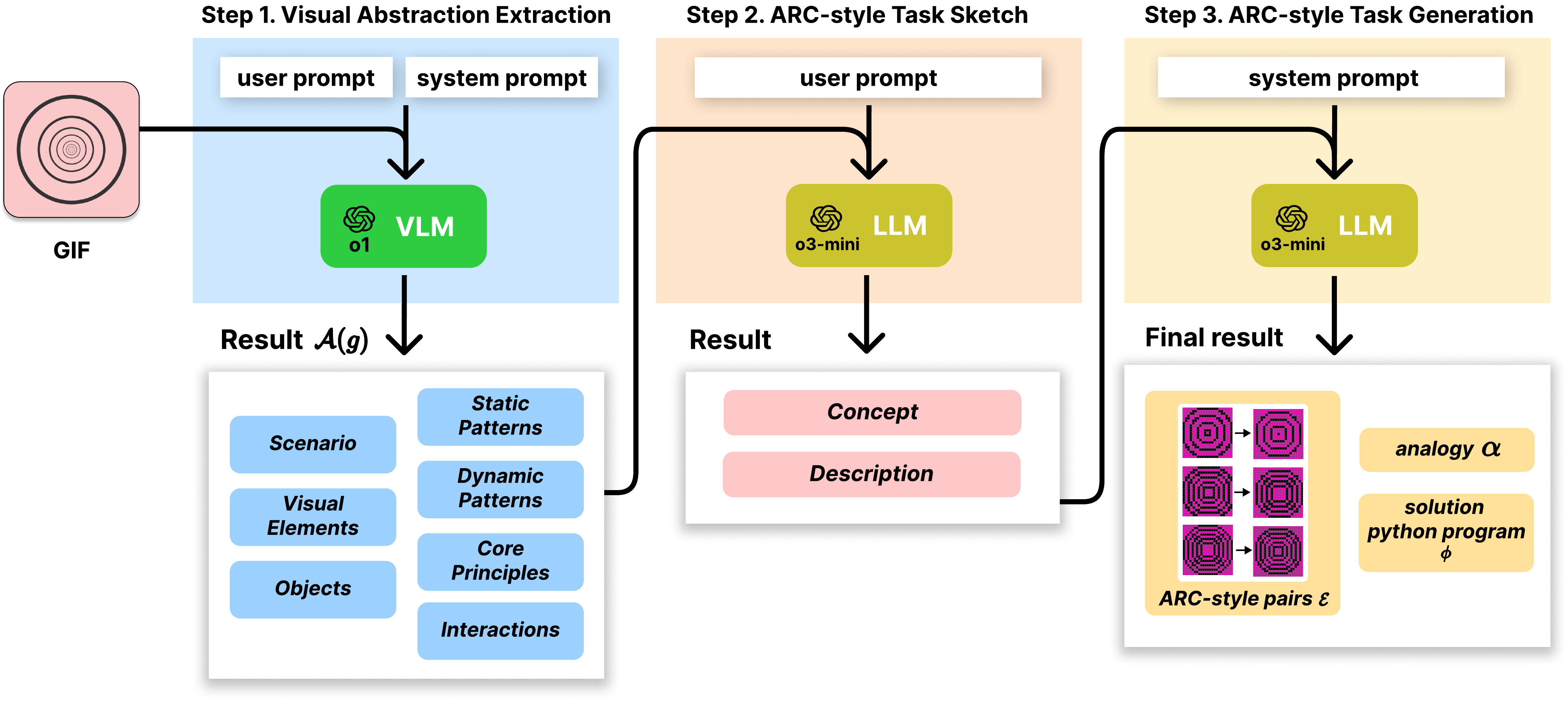}
    \caption{Illustration of GIFARC data synthesis pipeline that transforms a single GIF into a corresponding ARC-style task and supplementary data. In step 1, a vision language model (GPT o1) digests a GIF file and outputs a detailed text expression $\mathcal{A}(g)$ of the visual transformation implied in the GIF in a JSONL format.  In step 2, a large language model (GPT o3-mini) reads step 1 result JSONL and outputs a text sketch an ARC-style task. In step 3, a large language model (GPT o3-mini) reads the step 2 sketch result and generates an ARC-style task $\mathcal{E}$, an implied analogy $\alpha$, and the Python solution program $\phi$. The example prompts used in each step are reported in Appendix~\ref{sec:examples}.}
    \label{fig:main_framework}
\end{figure}

\paragraph{Pipeline summary.}
Our data synthesis pipeline is a three-stage process that transforms raw GIFs into fully executable, analogy-grounded ARC-style tasks (see Figure~\ref{fig:main_framework} for illustrative examples). 
\[
\text{GIF}\;\xrightarrow{}\;\underbrace{\text{Visual Abstraction}}_{\S~\ref{subsec:extract}}
\;\xrightarrow{}\;\underbrace{\text{Task Sketch}}_{\S~\ref{subsec:sketch}}
\;\xrightarrow{}\;\underbrace{\text{Executable ARC Task}}_{\S~\ref{subsec:task}}
\]

\begin{enumerate}[leftmargin=*,label=Step \arabic*.]
  \item A VLM (GPT o1) analyzes each GIF $g$ and outputs a structured JSON record that captures its scenario, objects, static/dynamic patterns, interactions, and core reasoning principles $\mathcal{A}(g)$.             
  \item A LLM (GPT o3-mini) converts the JSON abstraction into a concise ARC-style task sketch: a set of concepts with a natural-language description that implies the extracted patterns.  \looseness=-1 
  
  \item Conditioned on the sketch and a handful of retrieved ARC task examples, the LLM (GPT o3-mini) synthesizes Python code implementing \texttt{generate\_input} (stochastic examples) and \texttt{main} (deterministic transformation).  The output is a complete ARC-style task comprising (i)~an input-output demonstration set~$\mathcal{E}$, (ii)~the Python solution program~$\phi$, and (iii)~its analogy label~$\alpha$. \looseness=-1 
\end{enumerate}
Details of Step 1 appear in Section~\ref{subsec:extract}, Step 2 in Section~\ref{subsec:sketch}, and Step 3 in Section~\ref{subsec:task}. All LLM calls use Azure OpenAI endpoints. \looseness=-1

\subsection{Step 1. Extracting Visual Abstractions from GIFs}
\label{subsec:extract}

Our first stage converts a raw GIF into a structured, readable summary that captures the analogy-laden visual logic hidden in the clip. 
Given an animated GIF $g$ depicting a short visual phenomenon (e.g., snow piling up, light reflecting, pipes clogging), we wish to automatically extract the key elements $\mathcal{A}(g)$ that a human would use when reasoning by analogy: what the scene contains, which parts move or remain fixed, how objects interact, and which general principle explains the observed dynamics. The output, called a visual abstraction, serves as the input for the later stage.

\paragraph{GIF crawl.}
To maximize coverage of distinct visual regularities, we first assemble a visual transformation corpus of GIFs.
Using the GIPHY API, we query a curated tag list
\texttt{\{nature, geometry, pattern, relational\_pattern, sequence\}} where these types of clips often exhibit dynamic change.
After deduplication, we obtained total 10k GIFs.

\paragraph{Prompt for extracting visual abstractions.}
We then submit every GIF to the vision–language model GPT o1 under a prompt intentionally designed around two objectives: (i)~\emph{granularity} - to extract as many diverse visual analogies as possible from the GIF, (ii)~\emph{structure} – the reply must be valid JSON, and (ii)~\emph{coherence} – to ensure that each extracted visual analogy maintains alignment with the original GIF.
Full prompt used in this step is reported in Appendix~\ref{sec:full_prompt1}. The system message states the structural constraint (“return JSON with keys \{scenario ... interactions\} and use bullet lists only”). 
The user message enumerates the seven fields and attaches concrete bullet-level instructions. 

These seven fields in $\mathcal{A}(g)$ serve complementary purposes: (i) \texttt{scenario} describes the overall narrative of GIF images; (ii) \texttt{visual\_elements} stores the raw, static pixel facts; 
(iii) \texttt{objects} binds memorable names to entities so later prompts can reference them; 
(iv) \texttt{static\_patterns} and (v) \texttt{dynamic\_patterns} distinguish invariants and variants from transformations, respectively; 
(vi) \texttt{core\_principles} captures the very abstraction (e.g., gravity, reflection, occlusion, ...) that we ultimately hope an ARC solver will recall as an analogy; 
and (vii) \texttt{interactions} records explicit or implicit relations (e.g., “snowflake collides with ground”, “fluid flow is blocked”) among objects. 

This step is an extraction step to pull out intuitive analogies from GIF image. 
With this step, the GIFs can now be understood by text based LLMs in the later stages.
In addition, converting pixel data into a more lightweight text descriptions lets the pipeline to efficiently digest the analogy faster through the remaining synthesis stages without having to handle heavy image data along. 
Also, in case that GIF holds multiple visual analogies, explicitly pointing out which visual analogy to focus on lets the agents in later stages interpret the target analogy without possible confusion. 
Furthermore, this stage is critical in abiding the GIPHY API terms of right. 
By irreversiblly conceptualizing visual data into text description, we have generated a distinct visual analogy data pool that does not contain any explicit GIFs.

\subsection{Step 2. From Visual Abstraction to Task Sketch}
\label{subsec:sketch}
The visual abstraction \(\mathcal{A}(g)\) obtained in previous step is rich but verbose: it contains a lot of observable details of the GIF, many of which may be irrelevant for task design (e.g.,\ background textures, camera shakes). 
Before we can synthesize executable code, we therefore distill $\mathcal{A}(g)$ into a compact \emph{task sketch}
composed of two parts: ~(i) \textbf{Concepts:} 3–5 keyword phrases that name the core visual ideas (e.g., blocked\_flow, rotational\_symmetry, incremental\_accumulation, ...);~(ii) \textbf{Description:} a single paragraph that narrates how those concepts manifest on an ARC-style grid.

\paragraph{Prompt for task sketch.}
We obtain each sketch from a single completion of the text-only model GPT o1 using a three-block prompt: \textbf{Block 1} provides in-context demonstrations; 
we draw 75 concepts/description pairs at random from the 160 human-written example tasks released with the BARC corpus~\cite{li2024combining} and paste them.
Because these examples cover symmetry, counting, physical analogy, topological change, and more, they implicitly teach the model what “looks like” a solvable ARC-style task while leaving room for new content.
\textbf{Block 2} provides the seven fields of $\mathcal{A}(g)$ from the previous step in bullet form.
\textbf{Block 3} has a message that freezes the output format, exactly two comment headers for concepts and description.

We further instruct the model to reflect the full trajectory from initial to final state, maintaining semantic fidelity between the GIF dynamic behavior and the input-output structure of ARC-style grid.
Full prompt used in this step is reported in Appendix~\ref{sec:full_prompt2}.

This selection step is crucial in two aspects. 
First, we expect LLM to filter out abstractions that cannot be turned into
deterministic grid transformations (e.g., `clouds forming random shapes').  
Second, it forces the retained content into the exact format expected by our code-generation prompt via in-context learning, thereby lowering entropy and reducing failure cases downstream.

\subsection{Step 3. Generating Executable ARC-style Tasks}
\label{subsec:task}
The final stage transforms each task sketch produced in Step 2, into a fully executable ARC-style task.
The output comprises two parts: ~(i) \texttt{main}, A deterministic Python function implementing the
latent transformation $\mathcal{F}$ (rendered as Python code~$\phi$); ~(ii) \texttt{generate\_input}, A stochastic Python function that draws fresh grid examples from the same distribution to generate input grids.

From this output, input grids generated with \texttt{generate\_input} function are fed to the \texttt{main} function to produce corresponding output grids, thereby constructing a set of input-output grid pairs $\mathcal{E}$. The analogy $\alpha$ is copied from the task sketch created in Step 2, forming a ARC-style task $\mathcal{T}=(\mathcal{E},\alpha,\phi)$.

\paragraph{Prompt for generating executable ARC-style tasks.}
Our code-generation prompt inherits the overall structure introduced in BARC~\cite{li2024combining}, but differs in that the new task is grounded in the GIF-derived concepts and description produced in Step 2.
To tighten this grounding while preserving the coding style that has been proven effective in earlier literature, we again adopt an in-context learning strategy, along with retrieval augmentation.
Specifically, we embed both our description and the 160 human-authored example ARC-style tasks with \texttt{text-embedding-ada-002} model. 
The cosine similarity between sketch and examples is computed, and the top-4 semantically closest examples are pasted into the prompt before the new sketch.  
Each example already contains a working \texttt{generate\_input} / \texttt{main} pair, so these few-shot examples show the model how high-level concepts and narrative prose translate into concise Python code.  
The full prompt example for generating executable ARC tasks is reported in Appendix~\ref{sec:full_prompt3}.

\section{Dataset Analysis}
\label{sec:analysis}
To ground subsequent experiments, we first take a close look at the corpus itself.
We here provide our analysis on a sampled subset of the dataset. 
A comprehensive analysis and full statistics of the entire dataset are reported in Appendix.

\paragraph{Descriptive Statistics.}
Table~\ref{tab:basic_stats} summarizes coarse-grained corpus properties.
GIFARC currently contains 10,000 tasks, each accompanied by 103,357 train–test input/output pairs and an executable solution.
The average input grid is 420.404 cells with a standard deviation of 262.096, while target grids are at 1,125.484$\pm$2,894.980.
On average, a task uses 4.402 distinct colors (1.917 std), indicating that most tasks require multi-object reasoning or contain object with some complexity, rather than binary foreground/background segmentation. \looseness=-1

\begin{table}[h!]
  \centering
  \caption{Summary of GIFARC's descriptive statistics.}
  \label{tab:basic_stats}
  \begin{tabular}{lccc}
    \toprule
    & Count & Mean & Std. \\
    \midrule
    The number of tasks   & 10,000 & -- &-- \\
    Input–grid size ($h\!\times\!w$) & -- & 420.404 & 262.096\\
    Target–grid size ($h\!\times\!w$) & -- & 1,125.484 & 2,894.980\\
    The number of colors per task  & -- & 4.402 & 1.917\\
    \bottomrule
  \end{tabular}
\end{table}

\begin{figure}[t!]
\centering
    \includegraphics[width=\columnwidth]{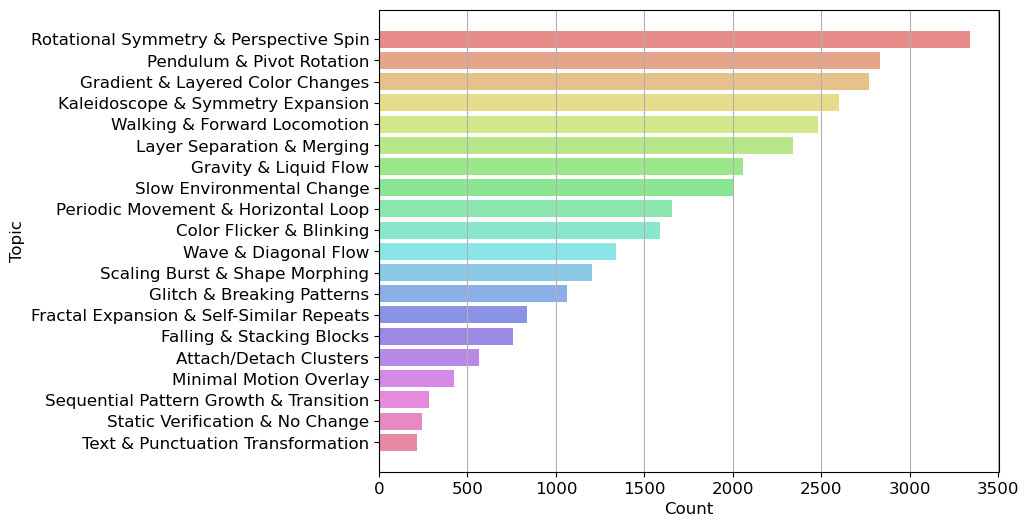}
    \caption{Histogram of task types occurrence in GIFARC.}
    \label{fig:task_type_histogram}
\end{figure}

\paragraph{Distribution of Task Types.}
To understand which semantic concepts and human-intuitive analogies dominate the dataset, we visualized the distribution of task types in our dataset.
We prompted GPT-4o to map each keyword in the task into one of 20 coarse types. 
Figure~\ref{fig:task_type_histogram} shows the histogram of task types based on the keyword. Note that in the initial stage of crawling raw GIFs, we sampled 500 GIFs each from 20 different GIF categories (e.g., nature, geometry, animation, ...). 
The distribution of the resultant task type has shifted from the initial GIF crawling category distribution. 
The results indicate that the problems generated from GIFs span a diverse range of task types, with the `Rotational Symmetry \& Perspective Spin' type being the most dominant.
Also, many problems contained a mixture of multiple task types.

\paragraph{Task Complexity.}
We assessed task complexity by measuring the complexity of the code used to generate GIFARC's grids. 
To this end, we report metrics such as the average number of lines of code (Lines of Code), the number of branching points in the code (Cyclomatic Complexity), the maximum depth of nested blocks in the code (Nesting Depth), and the number of unique operators used in the code (Unique Ops Count). These metrics make it easy to gauge how complex the processes required by GIFARC tasks are.

Table~\ref{tab:task_complexity} summarizes the code complexity metrics measured for GIFARC tasks.
The average number of lines of code is 12.12 with a standard deviation of 38.43, while the cyclomatic complexity averages 6.59 (3.66 std).
The maximum nesting depth in the code has a mean of 4.37 (1.52 std), and the number of unique operators averages 7.01 (0.18 std).
These metrics quantitatively demonstrate that GIFARC contains tasks with diverse levels of complexity.

\paragraph{Generation Fidelity.}
We measure the fidelity of our generation pipeline by evaluating the pass ratio at each stage.
A task is considered a failure if it fails to compile or does not produce outputs in the expected format.
Table~\ref{tab:pass_ratio} reports the success rates, showing that the overall pass ratio is consistently high across stages.

\begin{figure}[t!]
\centering
\begin{minipage}{0.49\linewidth}
  \centering
  \captionof{table}{Code Complexity of GIFARC Tasks}
  \label{tab:task_complexity}
  \begin{tabular}{lcc}
    \toprule
    & Mean & Std. \\ 
    \midrule
    Lines of Code  & 12.12 & 38.43 \\
    Cyclomatic Complexity & 6.59 & 3.66 \\
    Nesting Depth & 4.37 & 1.52 \\
    Unique Ops Count & 7.01 & 0.18 \\
    \bottomrule
  \end{tabular}
\end{minipage}%
\hfill
\begin{minipage}{0.47\linewidth}
  \centering
  \captionof{table}{Success rate of generation.}
  \label{tab:pass_ratio}
  \begin{tabular}{lc}
    \toprule
    & Rate (\%)\\
    \midrule
    Visual abstraction stage & 94.12 \\
    Task sketch stage & 80.54 \\
    Executable ARC task stage & 81.82 \\
    \bottomrule
  \end{tabular}
\end{minipage}%
\label{fig:difficulty_and_correctness}
\end{figure}


\section{Applications}
\label{sec:applications}

In this section, we conducted two experiments to examine: (1) how in-context learning with appropriately selected guiding examples influences the task-solving strategy of a LLM, and (2) how it affects the use of analogical information in GIFARC.

\subsection{Verification Changes in Reasoning Steps}
\label{subsec:verification}
In order to check the importance of each element in the dataset, GPT 4.1-mini was instructed to return the thought process while solving ARC-AGI-2 public evaluation dataset in context of sampled data of GIFARC. As described above, since ARC-AGI-2 dataset requires a more diverse approach than the predecessor, analogic approach in the task-solving process would be more clearly shown and used for the comparison of the reasoning step. 

GPT 4.1-mini was first learned in-context with 3 types of sampled GIFARC: 
\begin{enumerate} [leftmargin=*,label=(\roman*)]
    \item \textbf{full description}: 15 well-figuratively described data $(\mathcal{E},\alpha,\phi)$ chosen from GIFARC by repeated multi-turn refinement prompts to the LLM. 
    \item \textbf{description without analogy}: 15 data $(\mathcal{E},\alpha_{\text{flat}},\phi_{\text{flat}})$ from full description but gone through a `flattening' process where researchers replaced analogic terms in $\alpha$ and $\phi$ into their corresponding lower-level synonyms. 
    \item \textbf{description without analogy and without solution}: 15 data $(\mathcal{E},\alpha_{\text{flat}})$ from description without analogy but without solution $\phi_{\text{flat}}$.   
\end{enumerate}
Note that smaller numbered description contains richer analogies. These in-context learned models were then prompted to solve the tasks of ARC-AGI-2. In order to effectively check whether an analogic approach was used, models were instructed to specify two points in the result; first identify the underlying analogy in the given task, and implement it as a detailed solution.

\begin{figure}[t!]
\centering
    \includegraphics[width=\columnwidth]{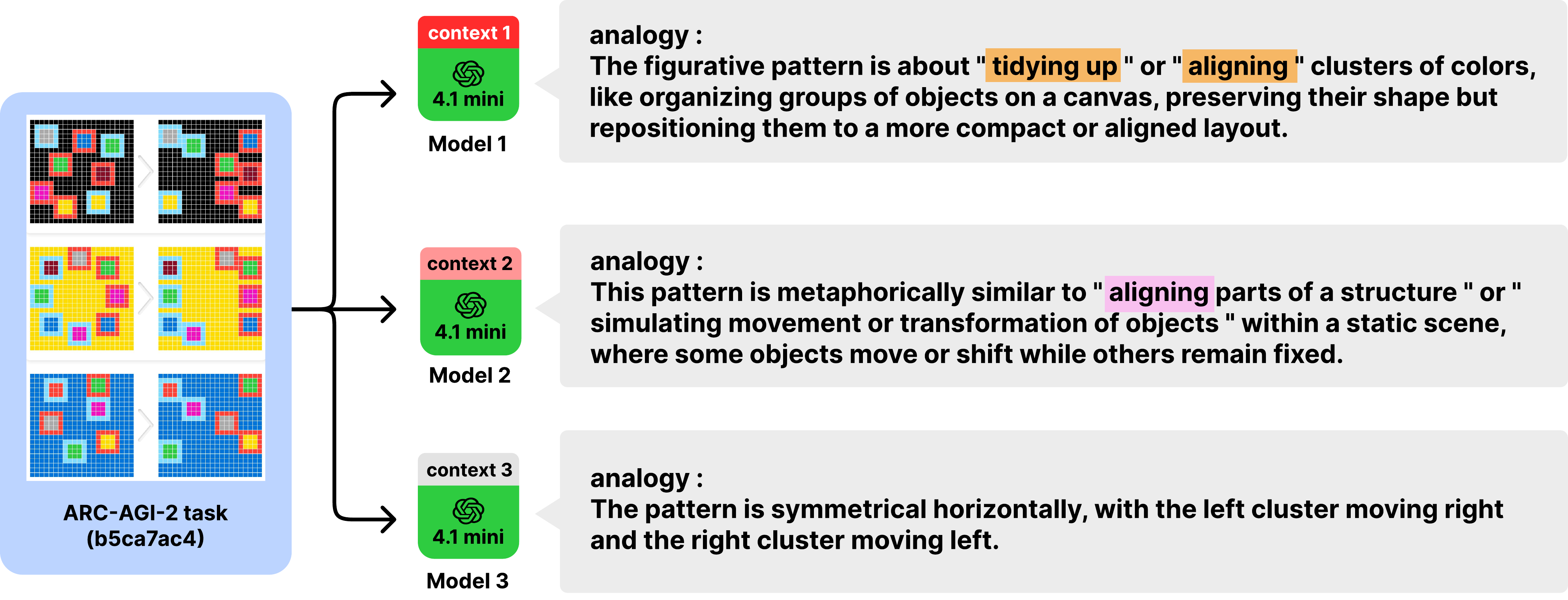}
    \caption{Experiment 1 case study: the LLMs learned in-context with richer analogy context result in more analogic interpretation of the ARC-AGI-2 task. Context 1, 2, 3 respectively stands for full description sample, description sample without analogy, description sample without analogy and without solution.}
    \label{fig:experiment_1_case_study}
\end{figure}

As a result of the solution comparison between models, we could confirm instances that analogies in GIFARC significantly helped identify the analogy of ARC-AGI-2 task. 
GPT 4.1-mini with full description expressed the analogy applied to the ARC-AGI-2 problem using an intuitive analogy. 
Figure~\ref{fig:experiment_1_case_study} depicts one of the well-performed result cases. 
While GPT 4.1-mini with description without analogy and without solution mainly described the task analogy with grid-level terms (symmetrical, horizontal, left, right), GPT 4.1-mini with full description was able to describe with words from several domains (such as `tidying up' and `organizing'). 
In particular, it was observed that GPT 4.1-mini perceived full description as not only as an `analogy phrase pool', but also recognized it as an `example of analogic approach' and could use external domain terminologies those are not present in context. 
This result shows that the analogic terms that were wiped during the flattening process are critical for the LLMs to get the grasp of analogic approach needed in solving ARC-style tasks. \looseness=-1

\subsection{Alignment between Context-Generated and Ground-Truth Analogies}
\label{subsec:alignment}
The second experiment was conducted to verify if LLMs with GIFARC-sampled context can successfully identify underlying analogy in ARC-style tasks. Here, we used full description and analogy-removed description, which we further removed the analogical expressions that remained in description (such as ``gravity'', ``spiral spin'', and ``flicker'') and replaced them with grid-level descriptions (such as ``move down '', ``rotate'', and ``color change''). Next, GPT 4.1-mini with full description and GPT 4.1-mini with analogy-removed description were instructed to find the analogy implied in the 12 ARC-style tasks. Along with two models, three humans (who knows the ARC well) also collected analogies implied in the 12 tasks. The outputs of two models and three humans were then compared with the ground-truth analogies of the task generated with GIFARC pipeline.

\begin{figure}[t!]
\centering
\subfloat[Similarity evaluation done by LLM]{
\includegraphics[width=0.42\linewidth]{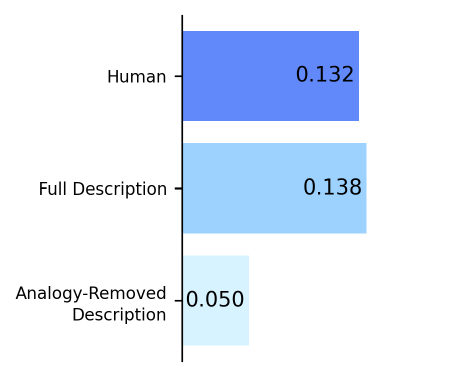}
\label{fig:similarity_by_LLM}
}
\subfloat[Embedding similarity]{
\includegraphics[width=0.42\linewidth]{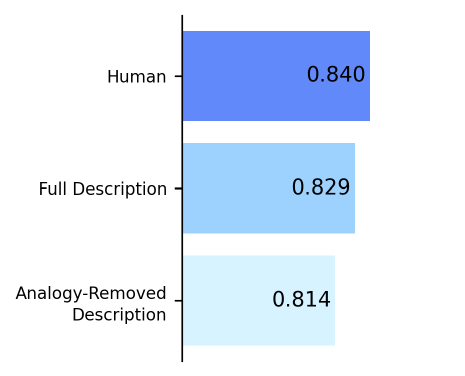}
\label{fig:embedding_similarity}}
\hspace{7mm}
\caption{Similarity analysis between task-implied analogy found by GIFARC-trained LLMs and its corresponding hand-crafted analogy. 
Figure~\ref{fig:similarity_by_LLM} illustrates that the analogy generated by GPT 4.1 mini with full description showed better similarity than that of GPT 4.1 mini with analogy-removed description by 0.087 pp, and embedded cosine similarity by 0.014 pp, as depicted in Figure~\ref{fig:embedding_similarity}.}
\label{fig:descriptive_statistics}
\end{figure}

We first instructed GPT o3-mini to evaluate semantic similarities by providing a one-shot guideline about measuring similarity. This evaluator LLM evaluated the semantic similarities between analogies generated by solvers (two in-context learned LLMs and three humans) and the ground truth analogy.
Figure~\ref{fig:similarity_by_LLM} depicts the similarity result done by LLM. It resulted that the output of GPT 4.1-mini with full description showed 0.137 similarity, while the output of GPT 4.1 mini with analogy-removed description only measured 0.050. 
Secondly, we embedded each outputs and compared the cosine similarity between. 
It resulted that the output of GPT 4.1-mini with full description showed 0.829 similarity, while the output of GPT 4.1 mini with analogy-removed description only scored 0.814.
This shows that context sample from GIFARC played a significant role in changing the task-solving approach of LLM to align with human-level analogic approach. Since GIFARC is a superset of the tested samples, it contains a much various types of analogies. Thus, this experiment shows the potential of GIFARC as the `analogic approach' guiding dataset.

\section{Conclusion}
\label{Sec:Conclusion}
We introduced GIFARC, a large-scale synthetic ARC-style dataset grounded in human-intuitive analogies extracted from GIF images. 
GIFARC framework narrows the human-AI reasoning gap by embedding explicit analogy labels and executable solutions that help language-vision models to map abstract grid transformations with everyday experiences. This approach has potential to improve AI-assisted education, scientific discovery, and decision-making support systems. No significant negative societal impacts are anticipated from this framework at present.
Empirical results on the original ARC benchmark confirm that both fine-tuning on GIFARC and leveraging analogy cues in the reasoning step boost solver accuracy. 
However, our data generation pipeline is inherently dependent on a single GIF for analogy extraction, which places a limitation on the diversity and scope of analogies.
In future work, we will scale GIFARC beyond motion-rich GIFs to videos and 3D simulations, extend evaluation to additional reasoning suites, and explore the fusion of multiple visual analogies to generate much more complex ARC tasks. 




\bibliographystyle{unsrt}
\bibliography{ARC_reference}

\newpage

\clearpage
\appendix

\section*{Appendix}
\section{Examples of Prompts}
\label{sec:examples}

\subsection{Full Prompt Example for Extracting Visual Abstractions}
\label{sec:full_prompt1}

The extraction of visual abstractions constitutes the first step in the data generation process. This process involves extracting comprehensive information from GIFs, including \texttt{scenario}, \texttt{visual\_elements}, \texttt{objects}, \texttt{static\_patterns}, \texttt{dynamic\_patterns}, \texttt{core\_principles}, and \texttt{interactions}. The specific content requirements for these seven types of information and the response format for the LLM are detailed in both the user prompt and system prompt below. The following content presents the complete specification of the aforementioned prompts.

\begin{tcolorbox}[breakable, colback=gray!5!white, colframe=gray!75!black, title=User Prompt: Extracting Visual Abstractions from GIFs]
\small{

Please analyze the given GIF by extracting core reasoning elements.  
For each section below, provide your response as a list of clearly separated items (not long-form paragraphs).  

\addvspace{8pt}

\textbf{[scenario]}
\begin{itemize}
    \item Describe the complete narrative of the GIF in 1-3 concise sentence(s).
    \item Include initial state, key changes, and final state.
    \item Include only observable facts and avoid speculation.
\end{itemize}

\addvspace{8pt}

\textbf{[visual\_elements]}  
\begin{itemize}
  \item Identify the key visual elements, objects, and actors in the GIF.
  \item List the major visual components in the scene (e.g., objects, colors, spatial arrangement, shapes).
\end{itemize}

\addvspace{8pt}

\textbf{[objects]}  
\begin{itemize}
  \item List all visual objects appearing in the GIF using this format:

    \begin{adjustwidth}{}{}
        \texttt{\{} \\
        \hspace*{1em}\texttt{"name": "object name",} \\
        \hspace*{1em}\texttt{"type": "explicit" or "implicit"} \\
        \texttt{\}}
    \end{adjustwidth}
 
  \item explicit = physical objects with clear boundaries  
  \item implicit = patterns or structures formed by multiple elements  
  \item Identify 2-8 key objects.
\end{itemize}

\addvspace{8pt}

\textbf{[static\_patterns]}  
\begin{itemize}
  \item Identify the elements or relationships that remain consistent throughout the GIF.
  \item Describe repeating spatial arrangements, consistent backgrounds, or fixed design features.
  \item List all patterns or objects that remain constant throughout the GIF.
\end{itemize}

\addvspace{8pt}

\textbf{[dynamic\_patterns]}  
\begin{itemize}
  \item Describe how the elements interact and change over time.
  \item Consider whether the changes occur gradually (step-by-step), abruptly, or through multi-stage transformations (e.g., directional shifts, scaling, rotation).
  \item List the distinct changes or interactions that occur over time.
\end{itemize}

\addvspace{8pt}

\textbf{[core\_principles]}  
\begin{itemize}
  \item Identify all general reasoning principles or mechanisms that explain \textbf{how and why} the dynamic elements change over time.
  \item Each principle should be \textbf{generalizable beyond the current GIF} and help abstract the reasoning structure behind the transformation.
  \item List one or more such principles that account for the observed dynamics.
  \item After listing them, \textbf{summarize the single most fundamental principle in one concise sentence.}
  \item Examples include: physical forces (e.g., gravity causes downward movement), goal-oriented behaviors (e.g., movement toward a target), causal chains (e.g., one event triggers another), symmetry-based transformations (e.g., reflection or alignment), and repetitive or cyclic patterns.
\end{itemize}

\addvspace{8pt}

\textbf{[interactions]}  
\begin{itemize}
  \item List object interactions using this format:

  \begin{adjustwidth}{}{}
    \texttt{\{} \\
    \hspace*{1em}\texttt{``objects\_involved'': [``object1'', ``object2''],} \\
    \hspace*{1em}\texttt{``interaction\_type'': ``clear/ambiguous/constraint'',} \\
    \hspace*{1em}\texttt{``interaction\_parameters'': [``parameter1'', ``parameter2'']} \\
    \texttt{\}}
  \end{adjustwidth}

  \item \texttt{interaction\_type} definitions:
  \begin{itemize}
    \item \textbf{clear}: distinct physical interactions (collisions, contact, etc.)
    \item \textbf{ambiguous}: indirect or unclear interactions
    \item \textbf{constraint}: interactions that establish limitations or boundaries
  \end{itemize}
  \item Record 2-6 key interactions, including only those directly observed in the GIF.
\end{itemize}
}
\end{tcolorbox}

\begin{tcolorbox}[breakable, colback=gray!5!white, colframe=gray!75!black, title=System Prompt: Extracting Visual Abstractions from GIFs]
\small{
You are an analysis assistant. Your role is to extract structured and reproducible information from a visual GIF using the five categories below.

\addvspace{6pt}

You \textbf{MUST} focus on observable phenomena, transformation structures, and etc.

\addvspace{6pt}

Do \textbf{NOT} invent unobservable events or make abstract generalizations. Describe only what can actually be seen in the GIF.

\addvspace{6pt}

For each category below, provide a plain list of items (one per line or bullet point).  
Avoid paragraph-style narration. Focus only on concrete, observable phenomena and transformation patterns.

\addvspace{10pt}

Return your response strictly in the following JSON format:

\begin{adjustwidth}{}{}
    \texttt{\{} \\
    \hspace*{1em}\texttt{``scenario'': ``<detailed description of the overall narrative>'',} \\
    \hspace*{1em}\texttt{``visual\_elements'': [} \\
    \hspace*{2em}\texttt{``<list of observable objects, colors, spatial arrangements, or notable visual traits>''} \\
    \hspace*{1em}\texttt{],} \\
    \hspace*{1em}\texttt{``objects'': [} \texttt{\{} \\
    \hspace*{3em}\texttt{``name'': ``<object name>'',} \\
    \hspace*{3em}\texttt{``type'': ``explicit/implicit''} \\
    \hspace*{2em}\texttt{\}} \texttt{],} \\
    \hspace*{1em}\texttt{``static\_patterns'': [} \\
    \hspace*{2em}\texttt{``<list of all objects or properties that remain unchanged throughout the GIF>''} \\
    \hspace*{1em}\texttt{],} \\
    \hspace*{1em}\texttt{``dynamic\_patterns'': [} \\
    \hspace*{2em}\texttt{``<list of all distinct transformations or movements that occur over time>''} \\
    \hspace*{1em}\texttt{],} \\
    \hspace*{1em}\texttt{``core\_principles'': [} \\
    \hspace*{2em}\texttt{``<list of general reasoning principles behind the transformation (e.g., gravity causes vertical motion)>''} \\
    \hspace*{1em}\texttt{],} \\
    \hspace*{1em}\texttt{``interactions'': [} 
    \texttt{\{} \\
    \hspace*{3em}\texttt{``objects\_involved'': [``<object1>'', ``<object2>''],} \\
    \hspace*{3em}\texttt{``interaction\_type'': ``clear/ambiguous/constraint'',} \\
    \hspace*{3em}\texttt{``interaction\_parameters'': [``<parameter1>'', ``<parameter2>'']} \\
    \hspace*{2em}\texttt{\}} \texttt{]} \\
\texttt{\}}
\end{adjustwidth}
}
\end{tcolorbox}

\newpage 

\subsection{Full Prompt Example for Task Sketch}
\label{sec:full_prompt2}

The task sketch represents the second step in the data generation process. The objective of this step is to design ARC-style tasks by dividing them into \texttt{concepts} and \texttt{description} components, utilizing the seven types of information extracted in Step 1. The user prompt provides the detailed specifications for this process.

\begin{tcolorbox}[breakable, colback=gray!5!white, colframe=gray!75!black, title=User Prompt: Task Sketch]

You've generated these on previous requests:

\texttt{\{examples\}}

\addvspace{8pt}

Based on your previous GIF analyses, I'd like you to create a new concept that incorporates the key elements from this analysis:

\texttt{scenario: \{scenario\}} \\
\texttt{objects: \{objects\}} \\
\texttt{static\_patterns: \{static\_patterns\}} \\
\texttt{dynamic\_patterns: \{dynamic\_patterns\}} \\
\texttt{interactions: \{interactions\}} \\
\texttt{core\_principles: \{core\_principles\}}

\addvspace{8pt}

Brainstorm one more.

\addvspace{8pt}

If the above informations involve gradual changes over time or a process of reaching a clear goal state, the input and output grids should not merely depict a short-term or single-frame transition. Instead, they should be designed to capture the entire transformation process from the initial state to the final state.

\addvspace{8pt}

When temporal progression or cumulative change is central to the analogy, construct the task code so that this flow and its outcome are clearly reflected through the difference between input and output.

\addvspace{8pt}

In this case, please create concepts and descriptions that effectively incorporate the above visual elements, static patterns, dynamic patterns, and core principles, following the format below, while referencing previous examples:

\begin{adjustwidth}{}{}
    \texttt{\# concepts:} \\
    \texttt{\# <concepts in your new generation>} \\[0.5em]
    \texttt{\# description:} \\
    \texttt{\# <description of your new generation>}
\end{adjustwidth}

\end{tcolorbox}

\newpage

\subsection{Full Prompt Example for Generating Executable ARC Tasks Version 1}
\label{sec:full_prompt3}

We generated ARC-style tasks using the analogies extracted from Step 1 and Step 2, employing two versions of prompts. Version 1 utilizes original BARC seeds that were used to generate input and output grid pairs. The following prompt contains the detailed explanation for Step 3 of Version 1. 

\begin{tcolorbox}[breakable, colback=gray!5!white, colframe=gray!75!black, title=User Prompt: Puzzle Implementation from Description (Version 1)]

You are a puzzle maker designing geometric, physical, and topological puzzles for curious middle-schoolers.

\addvspace{8pt}

Each puzzle consists of uncovering a deterministic rule, pattern, procedure, algorithm, or transformation law that maps inputs to outputs.
Both the inputs and outputs are 2D grids of colored pixels. There are 10 colors, but the order of the colors is never relevant to the puzzle.

\addvspace{8pt}

The middle schoolers are trying to discover this deterministic transformation, which can be implemented as a Python function called \texttt{main}.
Designing a puzzle involves also creating example inputs, which can be implemented as a Python function called \texttt{generate\_input}. Unlike \texttt{main}, the \texttt{generate\_input} function should be stochastic, so that every time you run it, you get another good example of what the transformation can be applied to.

\addvspace{8pt}

Here is a overview of the puzzle you are designing:

\addvspace{6pt}
\texttt{\{description\}}

\addvspace{8pt}

Please implement the puzzle by writing code containing the \texttt{generate\_input} and \texttt{main} functions. Use the following standard library (\texttt{common.py}):

\addvspace{6pt}
\texttt{\{common\_lib\}}

\addvspace{8pt}

Here are some examples from puzzles with similar descriptions to show you how to use functions in \texttt{common.py}:

\addvspace{6pt}
\texttt{\{examples\}}

\addvspace{8pt}

Your task is to implement the puzzle, following these steps:

\addvspace{6pt}

1. Inspect the example puzzle implementations, making note of the functions used and the physical/geometric/topological/logical details  \\
2. Inspect the new puzzle's description  \\
3. Brainstorm a possible implementation for the new puzzle  \\
4. Generate a code block formatted like the earlier examples with a comment starting \texttt{\# concepts:} listing the concepts and \texttt{\# description:} describing the inputs and transformation from the given description.

\addvspace{8pt}

When implementing code, please avoid using float type variables, numbers with decimal points, and the math library. These elements make it difficult to intuitively identify patterns between input and output grids. Instead, use only integer operations and basic arithmetic operators to clearly reveal the essence of the pattern. Please strictly follow this constraint when implementing your code.

\addvspace{8pt}

Also, When implementing code, please use the minimum number of lines possible. As code gets longer, its complexity increases, and if it becomes too detailed and complicated, people will find it difficult to intuitively understand the puzzle's rules just by looking at the input and output grids. Situations where one needs to analyze the code to understand the rules should be avoided. Please write concise and efficient code that clearly reveals the core pattern.

\addvspace{8pt}

Be sure to make the transformation \texttt{main} deterministic. Follow the description closely.

\end{tcolorbox}

\begin{tcolorbox}[breakable, colback=gray!5!white, colframe=gray!75!black, title=System Prompt: Generating Executable ARC Task (Version 1)]

You need to help me write code containing the \texttt{generate\_input} and \texttt{main} functions according to the given puzzle design.  
You must use the standard library (\texttt{common.py}). Create an appropriate puzzle following the given puzzle design concepts and description.

\addvspace{8pt}

When writing code, please use variable names that are meaningfully related to the core concepts of the problem.  
For example, if the problem involves snow falling phenomena, use variable names like \texttt{snowflake}, \texttt{precipitation}, \texttt{accumulation}, \texttt{gravity}, \texttt{obstacle}, etc.  
Specifically, when implementing the \texttt{generate\_input} function and \texttt{main} function, make sure each variable name is directly associated with the concepts in the problem.  
For instance, use \texttt{gravity\_strength} for a variable representing the intensity of gravity, and \texttt{obstacle\_positions} for storing the locations of obstacles — choose names that clearly reveal the role and meaning of each variable.

\addvspace{8pt}

Additionally, in the \texttt{generate\_input} function, please restrict the grid size to be between \texttt{1x1} and \texttt{30x30}.  
Do not create grids larger than \texttt{30x30}.  
Implement the \texttt{generate\_input} function that creates inputs appropriate for the problem and the \texttt{main} function that utilizes them while following these constraints.

\addvspace{8pt}

When implementing code, please avoid using float type variables, numbers with decimal points, and the math library.  
These elements make it difficult to intuitively identify patterns between input and output grids.  
Instead, use only integer operations and basic arithmetic operators to clearly reveal the essence of the pattern.  
Please strictly follow this constraint when implementing your code.

\addvspace{8pt}

Also, when implementing code, please use the minimum number of lines possible.  
As code gets longer, its complexity increases, and if it becomes too detailed and complicated, people will find it difficult to intuitively understand the puzzle rules just by looking at the input and output grids.  
Situations where one needs to analyze the code to understand the rules should be avoided.  
Please write concise and efficient code that clearly reveals the core pattern.

\addvspace{8pt}

When doing this, please output your solution following the JSON format specified below.

\begin{adjustwidth}{}{}
    \texttt{\{} \\
    \hspace*{1em}\texttt{``library'': ``<Write only the libraries used in the code. Ex. from common import* \textbackslash n import numpy as np \textbackslash n ....>'',} \\
    \hspace*{1em}\texttt{``main\_code'': ``<Write the main code part.>'',} \\
    \hspace*{1em}\texttt{``generate\_input\_code'': ``<Write the generate input code part.>'',} \\
    \hspace*{1em}\texttt{``total\_code'': ``<Write total code including libraries, main, generate\_input and given concepts and description>''} \\
    \texttt{\}}
\end{adjustwidth}

\end{tcolorbox}

\newpage

\section{  ARC Tasks Generate Pipeline Version 2 }

We renewed Version 2 of task generation pipeline for Step 3. This version bump was carried out because the BARC seeds are unsuitable for GIFARC task generation. In GIFARC dataset generated by the initial pipeline, we spotted several tasks which failed to express objects apprearing in GIF. This was a critical problem, as it resulted in tasks that are irrelevent to the original GIF. Thus, we added an additional stage that explicitly extracts objects that are presented in GIF which is Step 3-1. We extracted a list of codes that generates a grid representation of each object appearing in the GIF with Step 3-1. The object-generation code extracted in this additional step was human‑intuitive and well‑known.

\subsection{ Architecture of ARC Tasks Generate Pipeline (Version 2)  }

After completing Step 1, the detected objects are then categorized into explicit or implicit types. Objects that have clear physical quantities or concrete shapes were categorized as explicit objects (i.e., human, dog, text), whereas implicit objects are metaphorical entities formed by compositing multiple explicit objects (i.e., circle: peoples are sitting around in circles, big fish: school of small fish form a shape of big fish to protect themselves from shark.). We instructed LLM (GPT 4.1) to write each object generation code not only with rich comments and rich analogies with great details, but also with meanings aligned to its pixel representations. Next, we 1) paired each block of code with corresponding object name, 2) filtered out duplicates, and 3) embedded the names of the objects into vectors using all-MiniLM-L6-v2. 
Only explicit objects were stored in the vector database, as implicit objects are composites of explicit objects by definition and therefore cannot serve as fundamental units. With those preparations completed, these codes are additionally supplied to Step 3-1 and Step 3-2, along with the results of their former step results. 
During Step 3-2, we queried the embedded objects name to the vector database searching for the explicit object name and code with the highest cosine similarity score. We use the code as the seed to build the prompt, which then is used in task generation. An example seed is reported below in Listing below.

\begin{lstlisting}[language=Python, caption={Example seed of object generation code.}]
def generate_amplifier_bitmap(width=9, height=8, body_color=1, knob_color=2, speaker_color=0, border_color=0):
    
    # width: Number of horizontal pixels (minimum 3, maximum 30)
    # height: Number of vertical pixels (minimum 3, maximum 30)
    # body_color: Amplifier body color (values from 0 to 9)
    # knob_color: Knob color (values from 0 to 9)
    # speaker_color: Speaker area color (values from 0 to 9)
    # border_color: Border color (values from 0 to 9)
    
    width = max(3, min(30, width))
    height = max(3, min(30, height))
    bitmap = [[speaker_color for _ in range(width)] for _ in range(height)]
    # Draw border
    for x in range(width):
        bitmap[0][x] = border_color
        bitmap[height-1][x] = border_color
    for y in range(height):
        bitmap[y][0] = border_color
        bitmap[y][width-1] = border_color
    # Draw body (main amp area)
    for y in range(1, height-2):
        for x in range(1, width-1):
            bitmap[y][x] = body_color
    # Draw knobs (top row, spaced)
    knob_count = max(2, width//4)
    for i in range(knob_count):
        x = 1 + i * ((width-2)//(knob_count-1)) if knob_count>1 else width//2
        bitmap[1][x] = knob_color
    # Draw speaker (bottom row, center)
    speaker_width = max(1, width//5)
    speaker_start = (width - speaker_width)//2
    for x in range(speaker_start, speaker_start+speaker_width):
        bitmap[height-2][x] = speaker_color
    return bitmap
\end{lstlisting}

\subsection{ Example of Prompt (Version 2) }

In Version 2, we refined the seeds to convert the object data identified in Step 2 into a grid-based representation. The gridded objects express analogical relationships through explicit object interactions and transformations, serving as the core components of GIFARC task generation. The improved seeds thus enable the integration of a broader range of analogical information into ARC-style grids.

Below are full prompt examples for Version 2.
\begin{tcolorbox}[breakable, colback=gray!5!white, colframe=gray!75!black, title=User Prompt: Executable Object generation code (Version2)]
\texttt{\# My instruction:}\\
\hspace*{1em}\texttt{Look at the given \# Word and create a bitmap.}\\
\hspace*{1em}\texttt{The bitmap must use numbers 0--9, each number standing for a pixel color.}\\
\hspace*{1em}\texttt{Strictly follow the \# Format shown below.}\\[0.5em]

\texttt{\# Word:}\\
\hspace*{1em}\texttt{\{}\,\texttt{template}\,\texttt{\}}\\[0.5em]

\texttt{\# Format:}\\
 
\hspace*{1em}\texttt{\{}\\
\hspace*{2em}\texttt{"bitmap": [[]],}\\
\hspace*{2em}\texttt{"pixel\_meaning": \{}\\
\hspace*{3em}\texttt{"0": "Background", ...}\\
\hspace*{2em}\texttt{\},}\\
\hspace*{2em}\texttt{"parameter\_desc": \{}\\
\hspace*{3em}\texttt{"width":  "Bitmap width – determines the overall horizontal size.",}\\
\hspace*{3em}\texttt{"height": "Bitmap height – includes any frame, legs, etc.",}\\
\hspace*{3em}\texttt{...,}\\
\hspace*{2em}\texttt{\},}\\
\hspace*{2em}\texttt{"function\_code": "...",}\\
\hspace*{2em}\texttt{"sample\_execute\_code": "..."}\\
\hspace*{1em}\texttt{\}}\\
\texttt{}\\[0.5em]

\texttt{\# Additional requirements:}\\
\hspace*{1em}\texttt{- Provide parameterized Python generator code so the bitmap can be}\\
\hspace*{2.4em}\texttt{created with various parameters (width, height, etc.) for the given \# word.}\\
\hspace*{1em}\texttt{- The resulting JSON \textbf{must} include both \textit{pixel\_meaning} (number-to-meaning map)}\\
\hspace*{2.4em}\texttt{and \textit{parameter\_desc} (parameter explanations).}\\
\hspace*{1em}\texttt{- \textit{function\_code} and \textit{sample\_execute\_code} must be valid, runnable Python.}\\
\hspace*{1em}\texttt{- \textit{function\_code} must contain \textbf{only} the Python function.}\\
\hspace*{1em}\texttt{- In \textit{sample\_execute\_code}, each result variable must be named}\\
\hspace*{2.4em}\texttt{like \textbf{test\_bitmap\_(word)\_(idx)} — e.g.\ test\_bitmap\_bed\_1,}\\
\hspace*{2.4em}\texttt{test\_bitmap\_bed\_2, test\_bitmap\_bed\_3.}\\
\hspace*{1em}\texttt{- Also show example outputs for bitmap sizes from 3$\times$3 up to 30$\times$30.}\\
\hspace*{1em}\texttt{- Neither width nor height may ever exceed 30.}\\
\hspace*{1em}\texttt{- Ensure the code contains no infinite loops.}\\
\hspace*{1em}\texttt{- Pixel color mapping is fixed as follows:}\\
\hspace*{2.4em}\texttt{0: "Black", 1: "Blue", 2: "Red", 3: "Green", 4: "Yellow",}\\
\hspace*{2.4em}\texttt{5: "Grey", 6: "Pink", 7: "Orange", 8: "Teal", 9: "Maroon".}\\
\end{tcolorbox}

\begin{tcolorbox}[breakable, colback=gray!5!white, colframe=gray!75!black, title=User Prompt: Generating Executable ARC Task (Version 2)]
\begin{adjustwidth}{}{}
\texttt{\# My Insturction}\\
The object generation code in the following \#materials creates a bitmap of the object, mentioned in the function name and returns it.  By combining these objects, provide code that implements the interactions and transformations described in \#Interaction and Change, with extensive comments throughout.  The return value must be JSON, and both input and output must contain detailed comments. \\ [0.8em]

\texttt{\# Requirements}\\
\hspace*{1em}\texttt{1.\ At least 2 pairs of example problems (input/output)}\\
\hspace*{1em}\texttt{2.\ Use of predictable transformation laws or analogies}\\
\hspace*{1em}\texttt{3.\ Clear definition of each object's role, interaction, and rule}\\
\hspace*{1em}\texttt{4.\ Emphasis on fun/creativity and adjustable difficulty}\\
\hspace*{1em}\texttt{5.\ Complex and creative Puzzle}\\
\hspace*{1em}\texttt{6.\ \textit{input\_bitmap\_generation\_code} must be executable Python code, when it generate a bitmap, bitmap should be contained in the parameter which has formatted name,  \textit{(short descrition, in 10 char)\_input\_bitmap\_(idx)} is the format you should keep, However the  name of bitmap should start with alphabet so that it keeps proper python variable name}\\
\hspace*{1em}\texttt{7.\ \textit{solution\_code} must be executable Python}\\[0.8em]
\texttt{\# Creating an ARC-Style Puzzle Problem}\\
\hspace*{1em}\texttt{$\rightarrow$ At least 2 pairs of example input--output}\\
\hspace*{1em}\texttt{$\rightarrow$ There is a pattern/rule}\\
\hspace*{1em}\texttt{$\rightarrow$ The human/AI infers the rule and solves the problem}\\[0.8em]
\texttt{\# Detailed Objectives}\\
\hspace*{1em}\texttt{You \emph{must} use the provided functions to create objects and generate}\\
\hspace*{1em}\texttt{bitmaps with them.}\\[0.3em]
\hspace*{1em} \texttt{\{seed\}}\\

Using this, create at least 2--3 pairs of input--output bitmap examples and a puzzle problem with a consistent rule.  The rule must be something humans are meant to infer! \\ [0.8em]

\texttt{\# Example Result Format}\\
 \hspace*{1em}\texttt{\{}\\
\hspace*{2em}\texttt{"input\_bitmap\_generation\_code": "... \#( this must use object generate function )",}\\
\hspace*{2em}\texttt{"used\_concept": "...",}\\
\hspace*{2em}\texttt{"solution\_code": "def main(input): ..."}\\
\hspace*{1em}\texttt{\}}\\[0.8em]
(If possible) Answer generation code, Do not give me a bitmap, just give me the code to generate it. Input bitmap generation code must use object generate function to generate input bitmap.\\[1.0em]
\texttt{\# Problem Story}\\
\hspace*{1em}\texttt{\{description\}}\\[0.8em]
\texttt{\# Required Information}\\
\hspace*{1em}\texttt{\{story\}}\\
(e.g.: triangle + pillar, pattern + triangle, etc.)
Output rule: (e.g.: change color of overlapping area, only certain pattern positions change, etc.)
Bitmap size: (between 10x10 and 30x30)

\end{adjustwidth}

\end{tcolorbox}

\newpage

\section{Implementation Details}
\label{sec:implementation}
\subsection{Model hyperparameters}

In pipeline Version 1, we set the temperature to 0 in order to build deterministic datasets. In pipeline Version 2, we set the temperature hyperparameter to approximately 0.2 to produce more diverse outputs. \cite{peeperkorn2024temperaturecreativityparameterlarge}. Note that reasoning models like GPT o-series do not support temperature configuration, so this parameter was not set for those models. For more detailed hyperparameters, refer to Tables~\ref{tab:api_hyper_v1} and ~\ref{tab:api_hyper_v2}.

\begin{table}[h!]
  \centering
  \caption{OpenAI API Hyperparameters Verison 1}
  \label{tab:api_hyper_v1}
  \begin{tabular}{cccccc}
    \toprule
    Process Name & Role & Model Name & Max Tokens  & Top p &  Temperature \\
    \midrule
    Method Step 1 & Data Generator & GPT o1    & 2,048      & 1.0 &  -   \\
    Method Step 2 & Data Generator& GPT o3-mini    & 2,048    & 1.0   & -  \\
    Method Step 3 & Data Generator& GPT o3-mini    & 2,048    & 1.0   & -  \\
    Application & LLM Evaluator & GPT o3-mini    & 40,000    & 1.0   & - \\
    Application & LLM Solver & GPT-4.1-mini    & 32,768    & 1.0   & 0.0  \\
    \bottomrule
  \end{tabular}
\end{table}

\begin{table}[h!]
  \centering
  \caption{OpenAI API Hyperparameters Version 2}
  \label{tab:api_hyper_v2}
  \begin{tabular}{cccccc}
    \toprule
    Process Name & Role & Model Name & Max Tokens  & Top p &  Temperature \\
    \midrule
    Method Step 1 & Data Generator & GPT o1    & 2,048      & 1.0   &  -   \\
    Method Step 2 & Data Generator& GPT o3-mini    & 2,048    & 1.0   & -  \\
    Method Step 3-1 & Data Generator& GPT 4.1    & 2,048    & 1.0   & 0.2  \\
    Method Step 3-2 & Data Generator& GPT 4.1    & 2,048    & 1.0   & 0.2  \\
    \bottomrule
  \end{tabular}
\end{table}

\subsection{Data Generation}
GIFARC builds on the data generation process established by the prior work BARC~\cite{li2024combining}. The original BARC framework generated tasks using manually crafted functions called seeds. A key difference from BARC is that GIFARC employs GPT o1 to extract seven key pieces of information from GIFs in Step 2, which are then used to design tasks in Step 3. In Version 1, generation is based on the \texttt{Description}, whereas in Version 2, it utilizes \texttt{Pattern\_information} and \texttt{Object\_information}, leveraging the \texttt{object\_bitmap} seeds constructed in GIFARC. Based on the designed tasks, we use GPT o3-mini to generate code in Step 3 of Version 1 and use GPT 4.1 to generate code in Step 3-1 and Step 3-2 of Version 2 for both the \texttt{generate\_input} function that creates inputs and the \texttt{main} function that produces outputs. In Version 2, the input bitmap is generated using the given \texttt{object\_bitmap} functions based on the semantic meaning of pixel values, and the output bitmap is produced through the solution code. We create input and output grids and apply a filtering process using this generated code. 

 The filtering criteria consist of ten conditions adapted from BARC, as shown in Table~\ref{tab:filtering_conditions}. Input-output grid pairs that meet any of these conditions are filtered out. In GIFARC, creating tasks that incorporate analogies often results in complex code. Therefore, we set a maximum time limit of 300 seconds per task to provide sufficient processing time. Cases requiring more time are considered to be caught in infinite loops and are filtered out as timeout failures. This filtering process aims to enhance the quality of GIFARC-generated tasks.

\begin{table}[h!]
  \centering
  \caption{Filtering Conditions}
  \label{tab:filtering_conditions}
  \begin{tabular}{@{} p{4cm} @{\vspace{0.25em}} p{\dimexpr0.95\textwidth-4cm\relax} @{}}
    \toprule
    Filtering Condition & Meaning  \\ 
    \midrule
    Non-Deterministic & The output grid differ when the same transformation is applied multiple times to the same input.  \\
    Non Color Invariant & 1. Cases where the transformation function itself fails during the color permutation process.
    
    2. Either the permuted grid or the input grid is not well-formed. 
    
    3. The results differ even when only the colors are changed. \\ 
    Identity & The input and output are completely identical.   \\ 
    Non-Well Formed Output & The transformation result is not well-formed (i.e., not a 2D list with equal row lengths and integer values between 0-9).  \\ 
    Black Output & The output consists entirely of 0s (black pixels).   \\ 
    Timeout & Overall time limit exceeded.   \\ 
    Non-Well Formed Input & The input is not well-formed (i.e., not a 2D list with equal row lengths and integer values between 0-9).   \\ 
    Duplicate Input &The generated input is duplicated with existing ones.   \\ 
    \bottomrule
  \end{tabular}
\end{table}

Version 2 develops code to convert object information extracted in Step 2 into grid representations, thereby enriching tasks with deeper analogical information. These gridified objects serve as seed functions for generating GIFARC tasks where analogies manifest as explicit object interactions or transformations. Specifically, we generate code that can represent objects such as `dog', `tree', `human', and `pacman' extracted in Step 2 as grids through GPT 4.1. By using these codes as seeds, we enable the analogical changes and patterns in GIFARC tasks to emerge more clearly. During task generation, these object-level grid fragments are combined as seed functions to construct input grids containing various objects. The corresponding output grids are generated by applying inter-object interactions and transformations to these input grids. Unlike the BARC seeds used in Version 1, Version 2 seeds are semantically structured, generating tasks with clearer and more interpretable analogical patterns. Finally, we apply the same 300-second timeout and filtering pipeline as in Version 1 to ensure that only efficiently and reliably executable tasks are retained.
%


\subsection{Application}
We conducted experiments on Section~\ref{subsec:verification} which is ``Verification Changes in Reasoning Steps'' and Section~\ref{subsec:alignment} which is ``Alignment between Context-Generated and Ground-Truth Analogies''. For these experiments, we developed various prompt context conditions: full description, description without analogy, and description without analogy and without solution.

The full description was generated using o3-mini to enhance the analogies in both the $\alpha$ and $\phi$ of the GIFARC dataset. To prevent the inclusion of irrelevant analogies, we performed manual refinement of the generated content. For the description without analogy, we utilized o3-mini to remove analogy-related information from the full description. Subsequently, we converted any remaining analogical elements in both $\alpha$ and $\phi$ to grid-level representations. The description without analogy and without solution used the same analogical framework as the description without analogy. Across all kinds of descriptions, the grid information remained consistent.

In our experiments examining Section~\ref{subsec:verification} and Section~\ref{subsec:alignment}, we employed both the full description and the analogy-removed description. The full description was identical to that used in the Section~\ref{subsec:verification} application. The analogy-removed description was a refined version of the description without analogy used in the Section~\ref{subsec:verification} experiment, where we completely eliminated any remaining analogical information by either removing it entirely or converting it to grid-level representations, thus ensuring the complete absence of analogical content.

\newpage

\section{Guidelines for human evaluators  assessing the given tasks}
\label{sec:guide_human}
In the section~\ref{sec:applications} ``Application'', we conducted human evaluation on 12 tasks. Three experts familiar with ARC were asked to describe the analogies for 12 GIFARC tasks. To obtain responses of sufficient quality for our experiment, we provided minimal guidelines. The following is a brief outline of the guidelines provided to the human evaluators.

\begin{tcolorbox}[breakable, colback=blue!5!white, colframe=blue!75!black, title=Guidelines for Human Evaluators]

Please write the pattern analogy in English sentences for the provided 13 tasks (input-output pairs), referring to the content below:
 \\ \\
1. What the provided input-output pairs are visualizing \\
2. What kind of analogy the pattern changing from input to output can be related to or expressed as \\
3. What the colors or object shapes shown in the input/output symbolize or represent \\
4. Please write the expected rules if you don't know the task's analogy. \\
Ex-1) - It appears to be spreading from the center. \\
Ex-2) - The movement seems to change randomly and the colors are changing. \\

\addvspace{10pt}

[Working Example] \\
\{Example task\} \\

\addvspace{10pt}

Analogy: In this task the grid represents a scene where small buds (colored gray: 5) transform into blooming flowers. The transformation simulates the process of blooming: each bud expands its presence by turning its neighboring cells (up, down, left, right) into petal cells (colored gray: 5) while the original bud cell becomes the highlight of the bloom (colored yellow: 4).  \\

\addvspace{10pt}

[GIFARC Task 1] \\
\{Input-Output Pairs\} 

\addvspace{10pt}

\vdots

\addvspace{10pt}

[GIFARC Task 12] \\
\{Input-Output Pairs\} \\ 

\end{tcolorbox}

Based on this guideline, participants described the analogies for the given GIFARC tasks. The information collected contained only content about the analogies, and no personal or sensitive information was gathered.

\newpage

\section{Example Visualizations of GIFARC}
\label{sec:Example_vis}

\begin{figure}[h!]
\centering
    \includegraphics[width=\columnwidth]{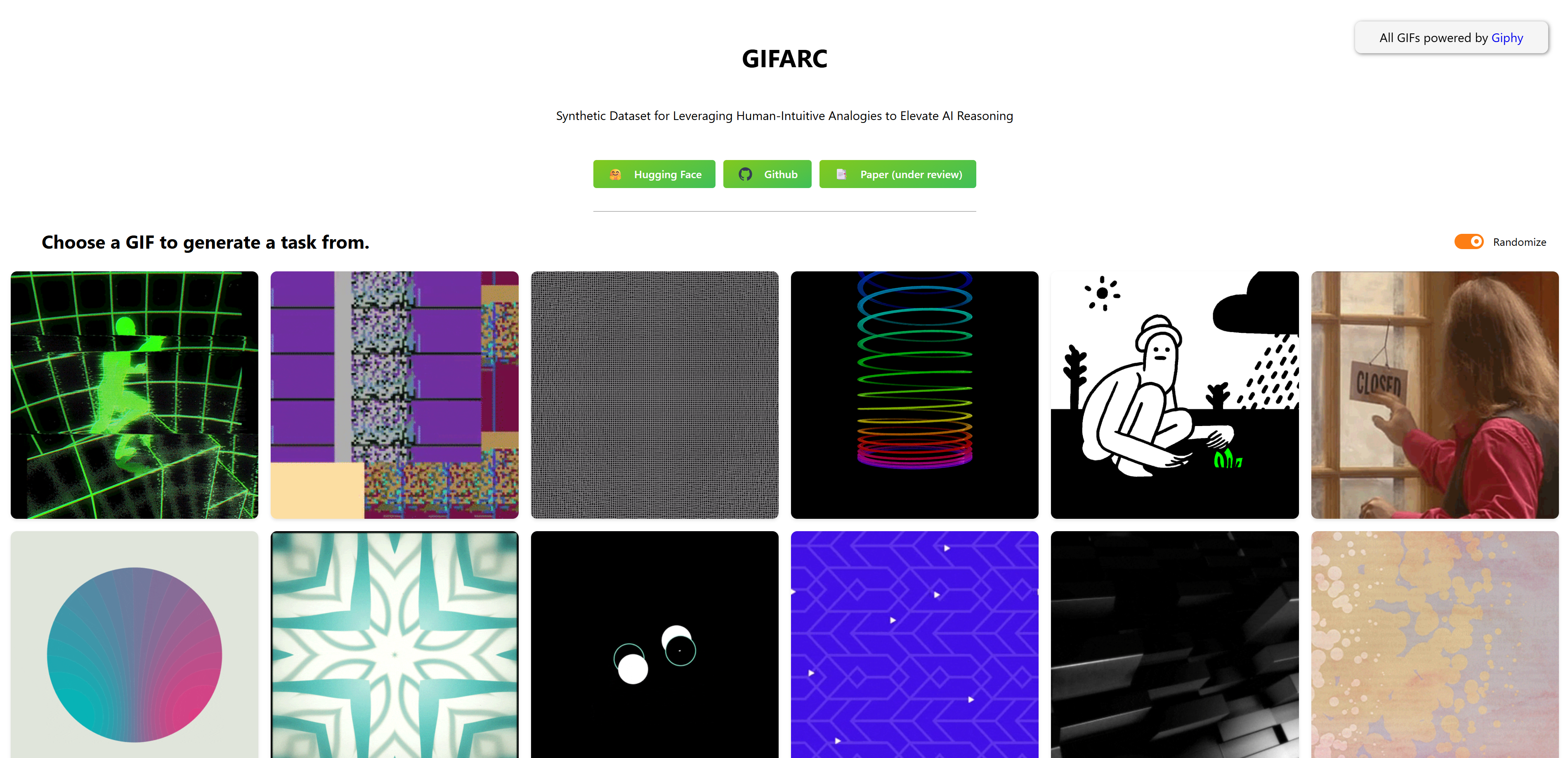}
    \caption{Landing Page of GIFARC Visualization website. It is publicly released at \url{https://gifarc.vercel.app/} }
    \label{fig:vis_website}
\end{figure}

All 10,000 tasks generated with GIFARC are visualized in \url{https://gifarc.vercel.app/}. 
Following are example visualizations of GIFARC-generated ARC-style tasks of each task type.

\subsection{Example Task of Type 1 - Rotational Symmetry \& Perspective Spin}
\begin{itemize}[leftmargin=*]
\begin{figure}[h!]
\centering
    \includegraphics[width=0.4\textwidth]{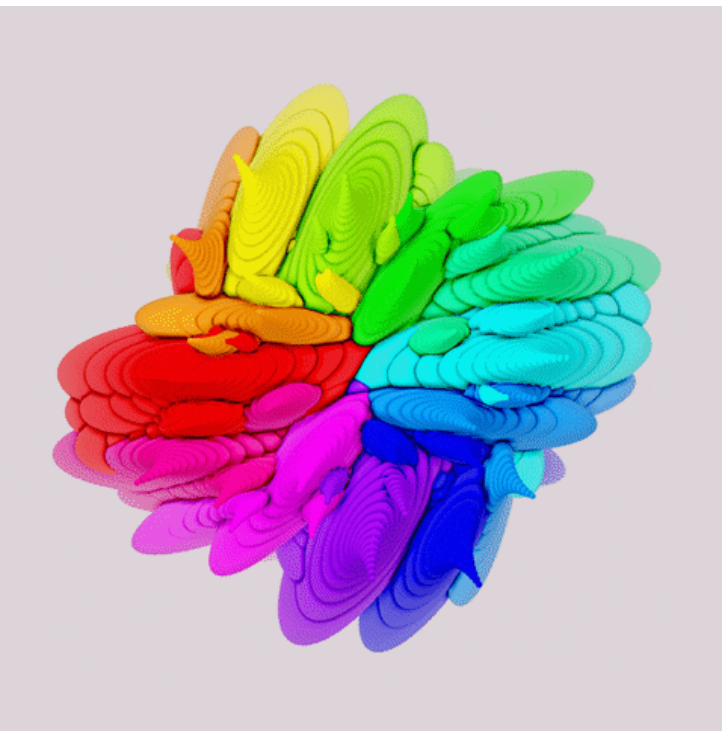}
    \caption{Snapshot of GIF used for generating task 962.}
    \label{fig:vis_type_gif_1}
\end{figure}

\item Concepts : fractal expansions, symmetrical pulsations, radial transformations, iterative growth
\item Description : In the input, you will see a sequence of grids (frames) on a black background.  Each grid depicts:  - A large circular region in the center, with multiple concentric rings and radial lines diverging outward.  - A stacked-curve fractal anchored at the bottom-left corner.  - A set of spire fractals anchored at the bottom-right corner.  As you move from one frame to the next in the input, these elements undergo iterative transformations:  - The radial lines repeatedly shift in thickness and visual intensity, while preserving rotational symmetry.  - The concentric rings in the circular region pulsate by alternately expanding and contracting.  - The stacked-curve fractal on the bottom-left changes its curve density, fractally adding or removing segments.  - The right-side spire fractals expand and contract in repeated vertical segments.  Your task is to replicate and apply these transformations for the entire sequence, and produce the final frame of the animation as the output:  1) Ensure the anchoring of the fractals at the bottom edge remains the same.  2) Preserve the radial symmetry of the central circular region and its common center point.  3) For each pulsation step in the input, magnify or contract the rings, lines, and fractals accordingly.  4) Continue until the final pulsation step is reached. That final state is your output grid.  The essential principle is that symmetry-based fractal transformations and repeated cyclical expansions/contractions produce the overall pulsating effect.  By reflecting each iterative change step by step, you reconstruct the final pulsating pattern visible at the end of the sequence.
\item Full web view is available at  \url{https://gifarc.vercel.app/task/962}.

\begin{figure}[h!]
\centering
    \includegraphics[width=\textwidth]{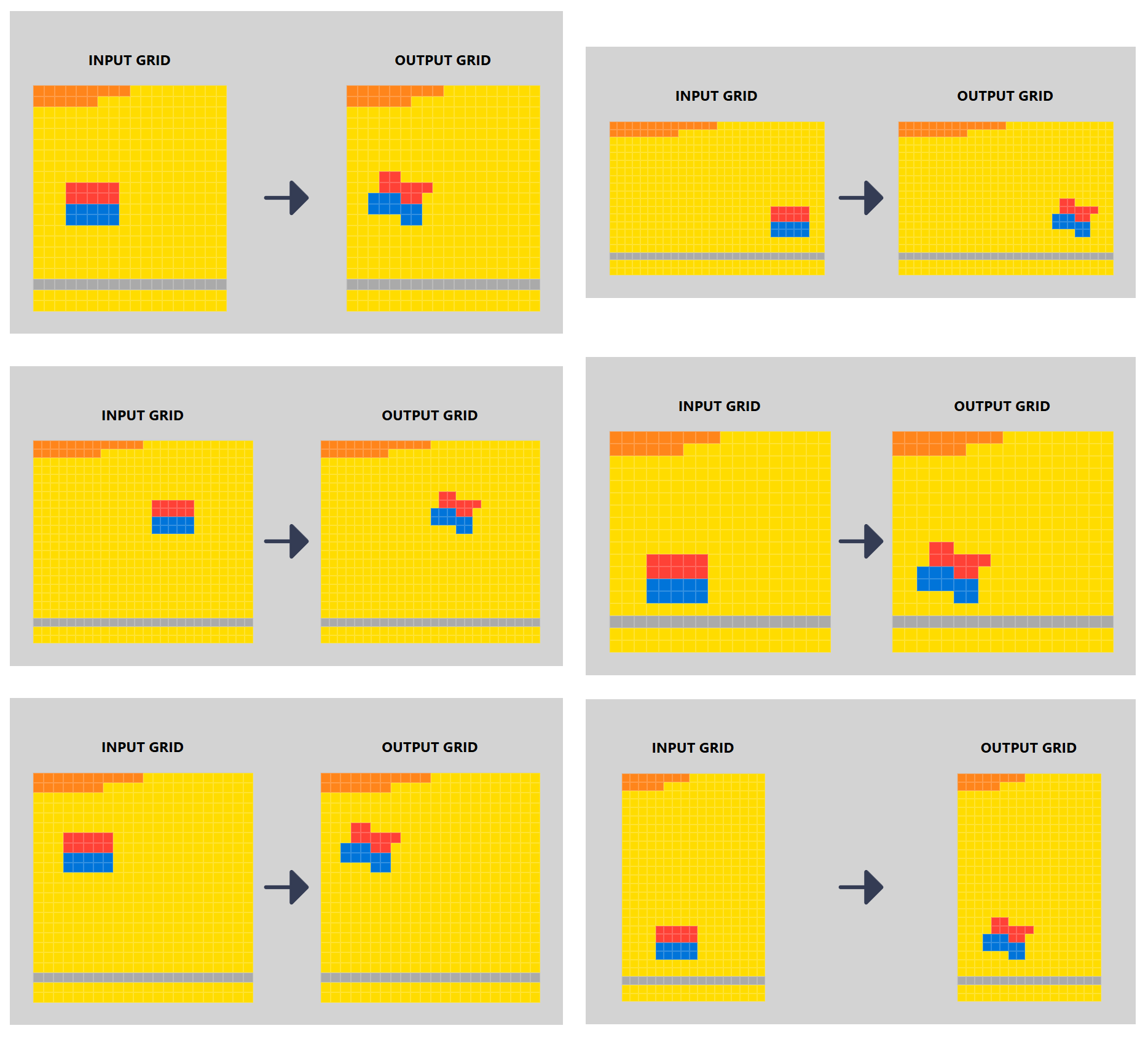}
    \caption{GIFARC-generated task 962.}
    \label{fig:vis_type_1}
\end{figure}

\end{itemize}

\newpage
\begin{lstlisting}[language=Python, caption={Python code solution for task 962.}]
from common import *
import numpy as np
import random

# concepts:
# rotational motion, center-of-mass shifting, multi-step transformation, static background

# description:
# The input grid shows a paved surface (Color.YELLOW) with static features: a fence (Color.GREY) drawn near the bottom and mountains (Color.ORANGE) at the top.
# A dynamic object-a wheelchair with its occupant-is represented by a 4x5 sprite: the top two rows are the occupant (Color.RED) and the bottom two rows the wheelchair (Color.BLUE).
# The wheelchair rotates around an axle (computed as the average position of the blue pixels) so that the occupant's mass shifts further off support.
# The transformation in main erases the original dynamic pixels and re-blits them using an integer-approximated clockwise rotation (about 22 degrees) around the axle, while leaving the static background unchanged.

def main(input_grid):
    out=input_grid.copy()
    dyn=[]
    for i in range(len(input_grid)):
        for j in range(len(input_grid[0])):
            if input_grid[i,j] in [Color.BLUE,Color.RED]:
                dyn.append((i,j,input_grid[i,j]))
                out[i,j]=Color.YELLOW
    if not dyn: return out
    ax_i=sum(p[0] for p in dyn)//len(dyn)
    ax_j=sum(p[1] for p in dyn)//len(dyn)
    cos_val, sin_val = 92,38
    for i,j,col in dyn:
        di=i-ax_i; dj=j-ax_j
        ni=ax_i+((cos_val*di+sin_val*dj)//100)
        nj=ax_j+((-sin_val*di+cos_val*dj)//100)
        if 0<=ni<len(out) and 0<=nj<len(out[0]): out[ni,nj]=col
    return out


def generate_input():
    H,W=random.randint(15,30),random.randint(15,30)
    grid=np.full((H,W),Color.YELLOW)
    for j in range(W): grid[H-3,j]=Color.GREY
    for i in range(2):
        for j in range(W//(2+i)):
            grid[i,j]=Color.ORANGE
    sprite=np.full((4,5),Color.BLACK)
    sprite[:2,:]=Color.RED; sprite[2:,:]=Color.BLUE
    x,y=random_free_location_for_sprite(grid,sprite,background= Color.YELLOW)
    blit_sprite(grid,sprite,x,y)
    return grid
\end{lstlisting}

\newpage
\newpage
\subsection{Example Task of Type 2 - Kaleidoscope \& Symmetry Expansion}
\begin{itemize}[leftmargin=*]
\item Concepts : kaleidoscopic transformation, mirrored folding, cyclical animation, vertical symmetry
\item Description : In this puzzle, the input is a multi-frame sequence depicting a pale-yellow corridor-like structure  with a central rectangular opening and mirrored arches aligned along a vertical axis of symmetry. Over several frames, the corridor's walls fold inward and outward in a smooth, kaleidoscopic motion,  always returning to the initial configuration. The output is the full cycle of transformations repeated in a loop,  preserving the corridor's pale-yellow hue and central frame while continuously reflecting the mirrored arch patterns  around the vertical axis. This cyclical approach showcases the corridor morphing into itself, completing  each transformation stage and then re-starting at the original symmetrical configuration.
\item Full web view is available at  \url{https://gifarc.vercel.app/task/7442}.

\begin{figure}[h!]
\centering
    \includegraphics[width=\columnwidth]{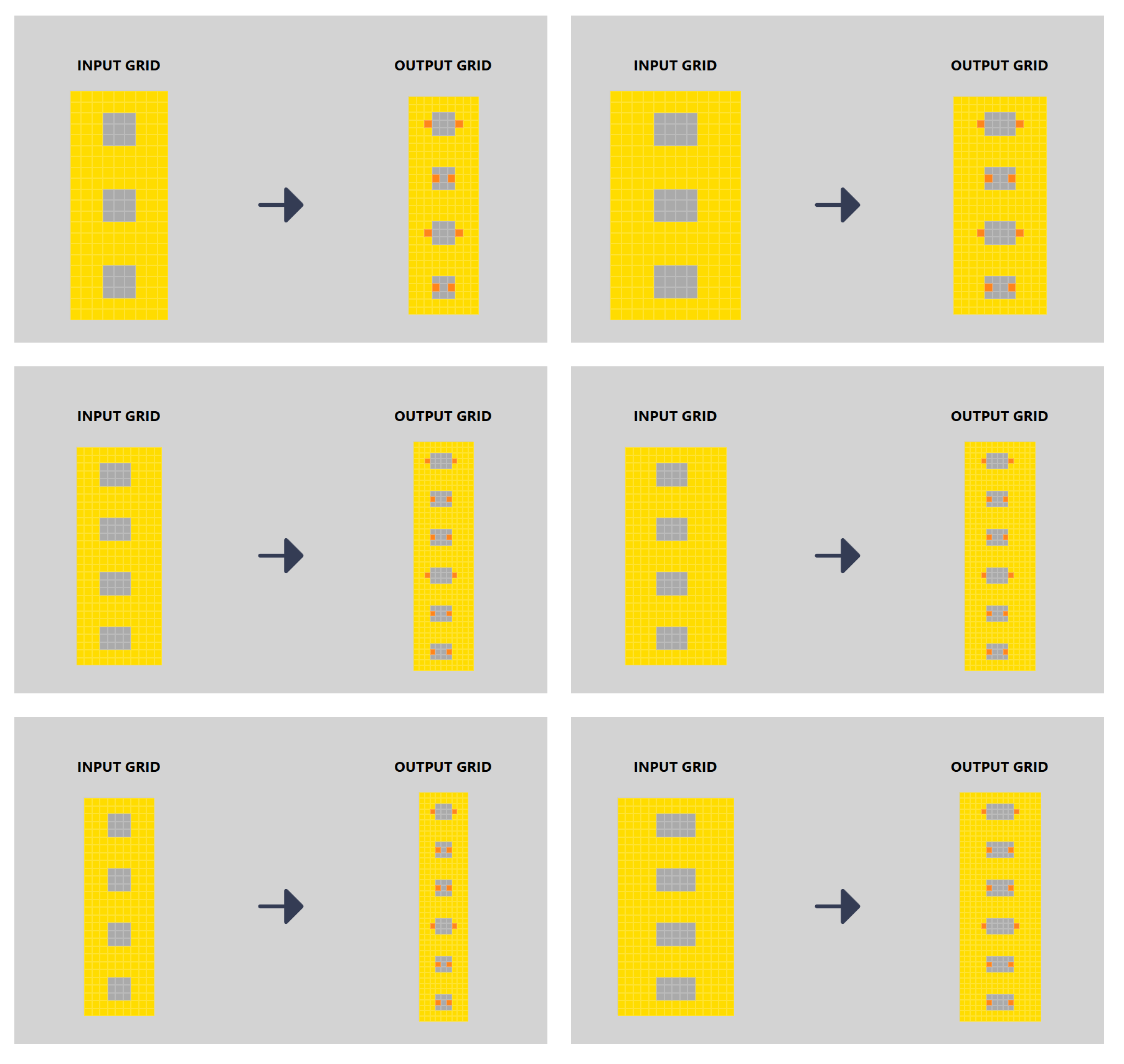}
    \caption{GIFARC-generated task 7442.}
    \label{fig:vis_type_2}
\end{figure}

\end{itemize}

\newpage
\subsection{Example Task of Type 3 - Pendulum \& Pivot Rotation}
\begin{itemize}[leftmargin=*]
\item Concepts : cyclical rotation, pivot anchoring, animation frames
\item Description : Inspired by previous examples that repeated or translated objects across multiple frames, in this puzzle the input is a single snapshot showing a blue curved shape on a black background. For the output, construct a multi-frame grid depicting the same shape completing a full 360-degree rotation around a central pivot. Each frame rotates the shape by a fixed angle (e.g., 15° increments), and the frames are arranged in a 3x3 or 4x4 grid until the shape returns to its original orientation.  The black background and the blue color remain consistent across frames, and the shape never goes outside the canvas boundary.  This cyclical rotation from start to finish, then repeating, is the puzzle’s core concept.
\item Full web view is available at  \url{https://gifarc.vercel.app/task/1037}.

\begin{figure}[h!]
\centering
    \includegraphics[width=\columnwidth]{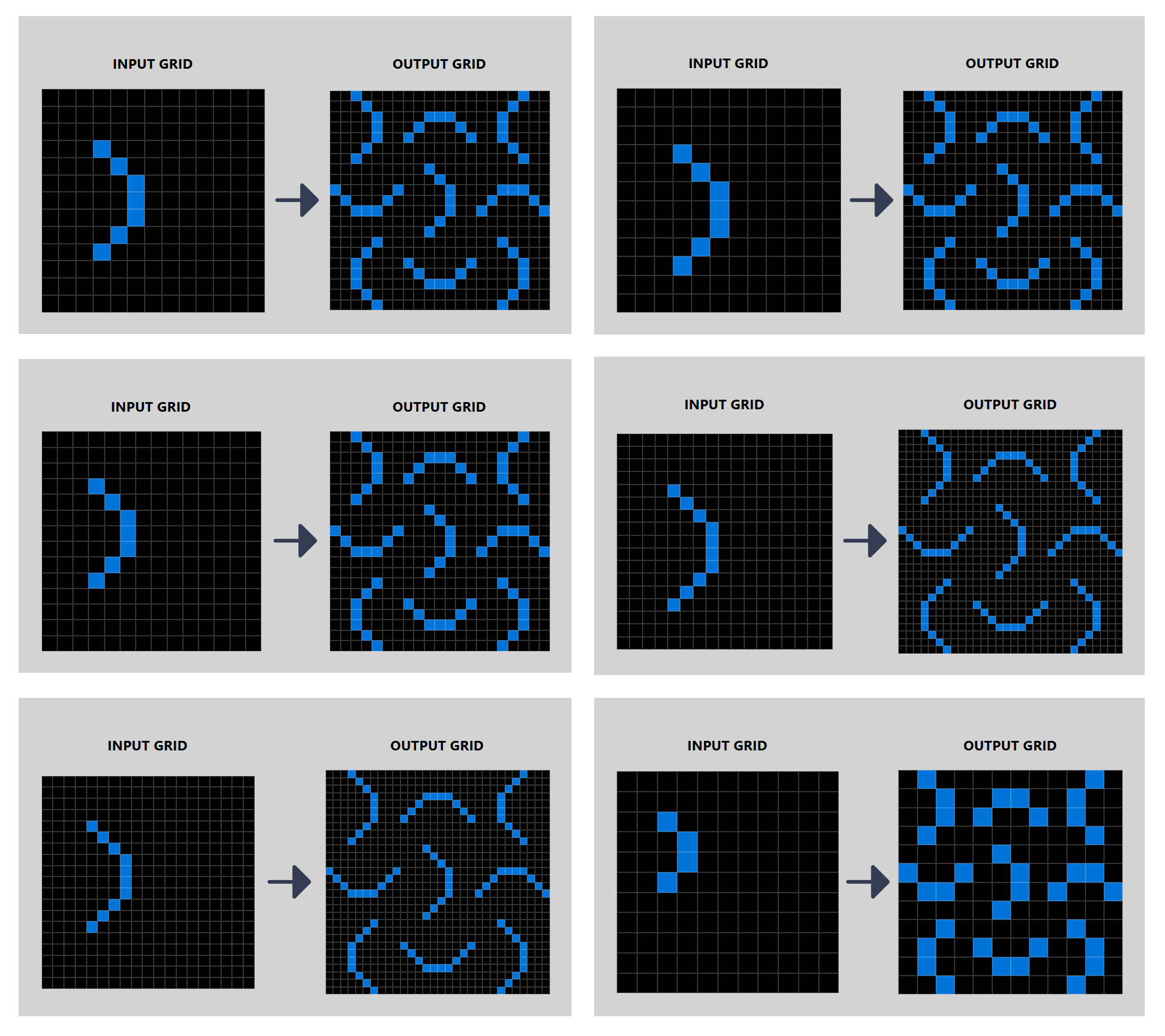}
    \caption{GIFARC-generated task 1037.}
    \label{fig:vis_type_3}
\end{figure}

\end{itemize}

\newpage
\subsection{Example Task of Type 4 - Walking \& Forward Locomotion}
\begin{itemize}[leftmargin=*]
\item Concepts : cyclical motion, repetition, background constraint
\item Description : In the input you will see an animation (or multiple frames) of pink human‑shaped figures in a light‑blue background.  White wavy lines span the background and shift subtly as the pink figures travel diagonally from the top‑left to the bottom‑right.  When a pink figure leaves the bottom‑right edge, it reappears at the top‑left edge at the same vertical position, making a continuous loop.  The translucent blue overlay patterns remain visually consistent on top of the background while the white lines gently oscillate as the figures pass. To produce the output, replicate this cyclical motion fully at least once, preserving continuity.  The final frame must line up so that if we start again from there, the pink figures continue repeating the same path through the light‑blue background in a seamless loop.
\item Full web view is available at  \url{https://gifarc.vercel.app/task/2767}.

\begin{figure}[h!]
\centering
    \includegraphics[width=\columnwidth]{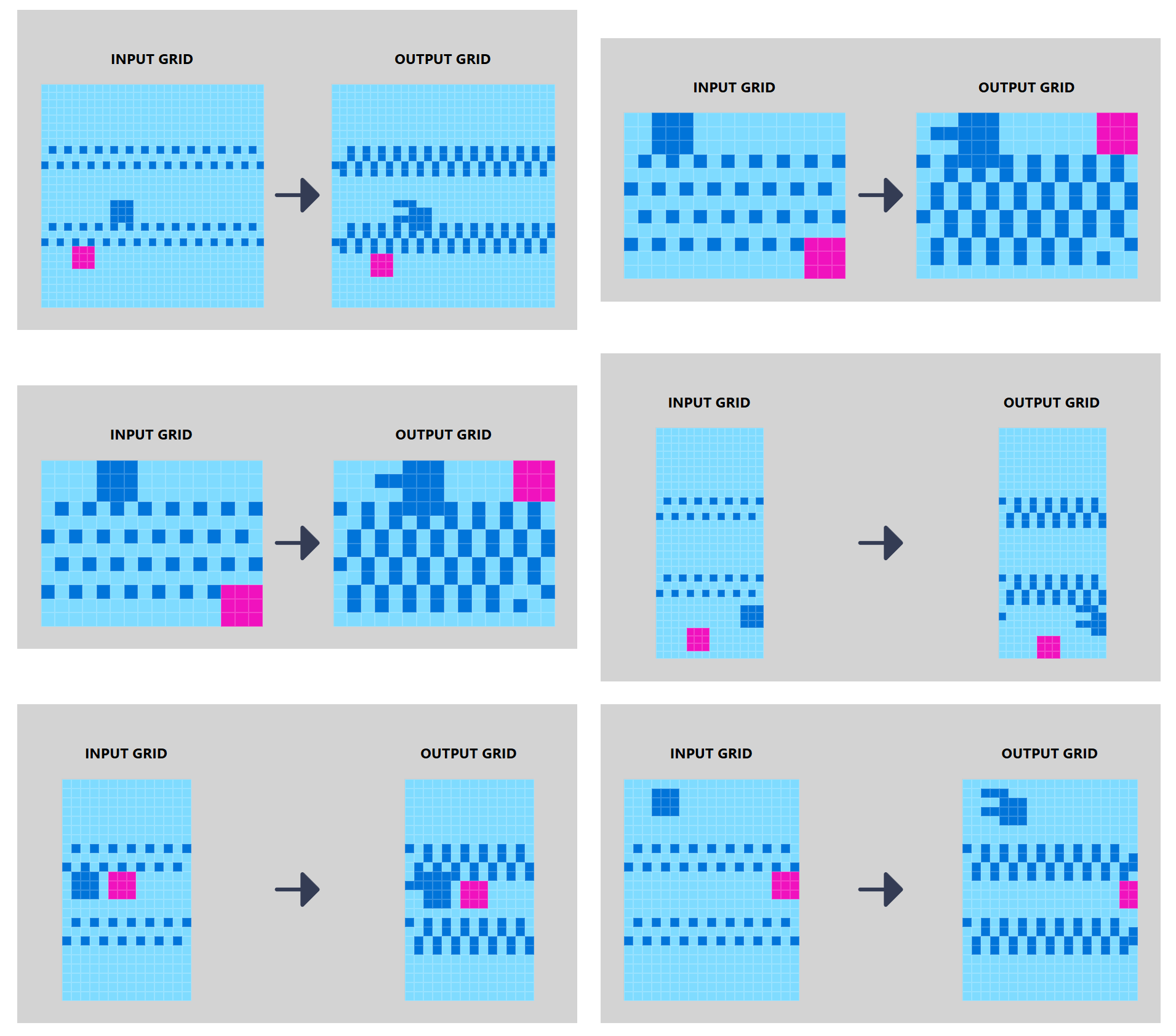}
    \caption{GIFARC-generated task 2767.}
    \label{fig:vis_type_4}
\end{figure}

\end{itemize}

\newpage
\subsection{Example Task of Type 5 - Falling \& Stacking Blocks}
\begin{itemize}[leftmargin=*]
\item Concepts : handheld device, screen area, gravity, falling blocks, stacking, looping animation
\item Description : In the input, you will see a pink background with a grey handheld gaming device in the center.  The device has a bright green rectangle acting as its screen, with a d-pad on the left and action buttons on the right. Initially, you will see Tetris-like blocks of various shapes hovering above the upper edge of this green screen area. To create the output, you must simulate the blocks falling downward into the screen, stacking on top of one another or on previously fallen blocks. Keep each block fully within the green screen boundary; if a falling block collides with another block or the bottom of the screen, it must stop. Once the screen is completely filled to the top, the stack disappears, effectively resetting the screen to empty and repeating the cycle. In other words, show the progression from “empty screen” → “stacked blocks” → “reset,” producing a looping effect of the falling Tetris-like blocks within the green screen.
\item Full web view is available at  \url{https://gifarc.vercel.app/task/556}.

\begin{figure}[h!]
\centering
    \includegraphics[width=\columnwidth]{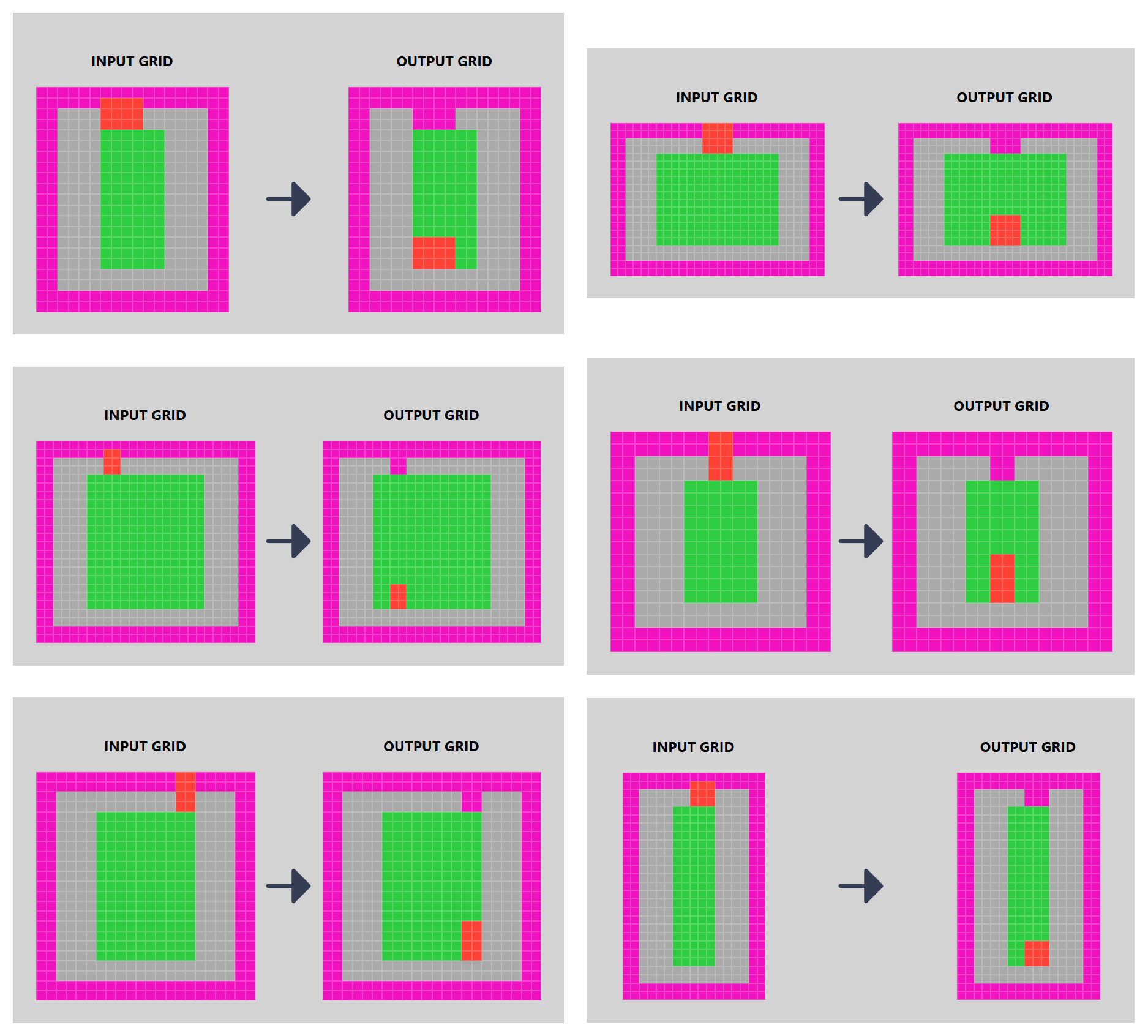}
    \caption{GIFARC-generated task 556.}
    \label{fig:vis_type_5}
\end{figure}

\end{itemize}

\newpage
\subsection{Example Task of Type 6 - Periodic Movement \& Horizontal Loop}
\begin{itemize}[leftmargin=*]
\item Concepts : cyclical motion, frame-based animation, horizontal repetition, background preservation
\item Description : In the input you will see several frames depicting a single brown triceratops, side-profile, walking across a beige background. The frames show a gradual shift in the triceratops’ position from left to right while its legs cycle through a walking gait.  To create the output, stack or tile these frames into a continuous strip (or series) that repeats seamlessly. The background stays beige and does not change between frames, and the triceratops remains the same brown color and orientation. The final output shows a horizontally looping animation strip where the last frame transitions smoothly back into the first, preserving the cyclical walking motion and the uniform background.
\item Full web view is available at  \url{https://gifarc.vercel.app/task/3690}.

\begin{figure}[h!]
\centering
    \includegraphics[width=\columnwidth]{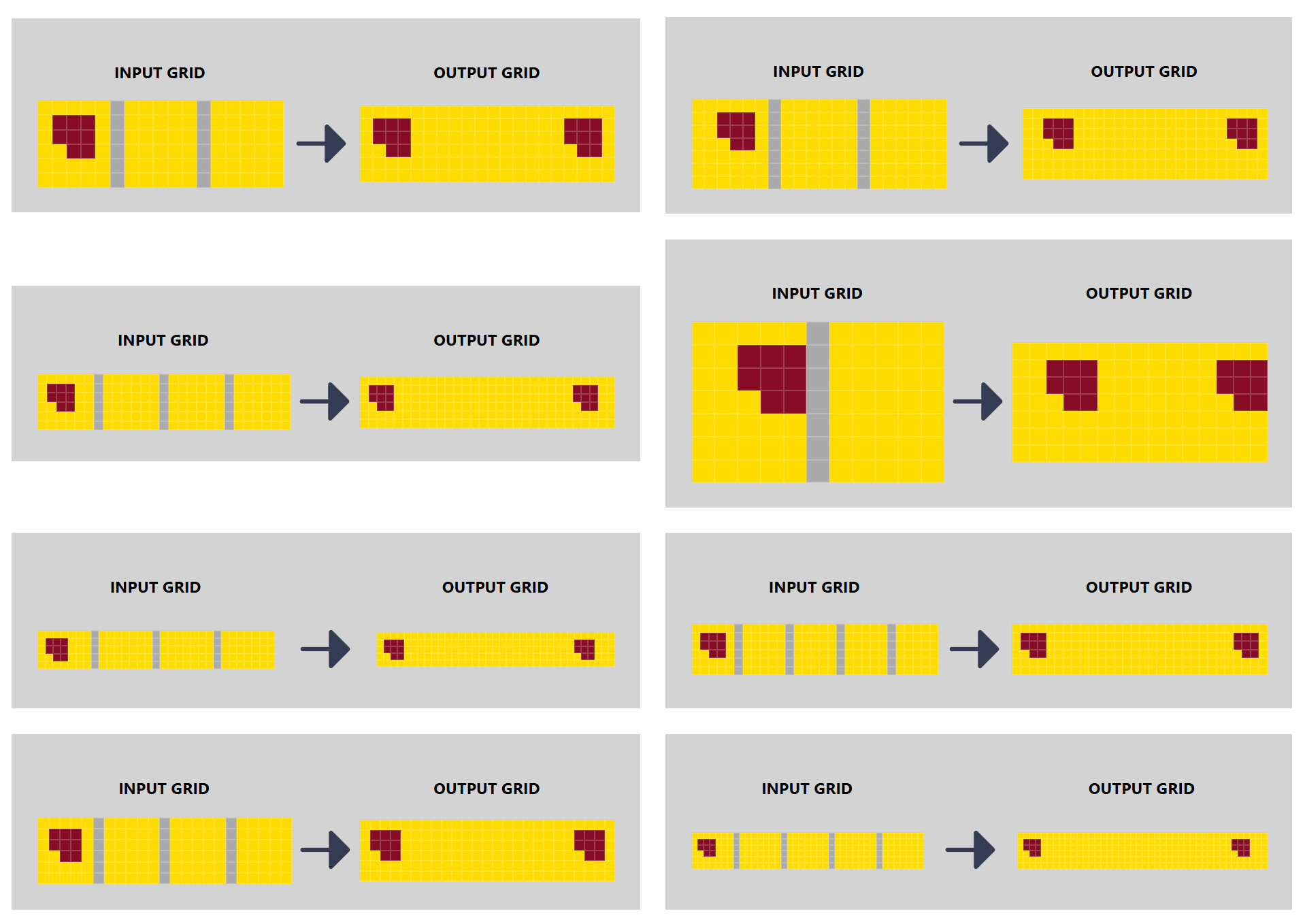}
    \caption{GIFARC-generated task 3690.}
    \label{fig:vis_type_6}
\end{figure}

\end{itemize}

\newpage
\subsection{Example Task of Type 7 - Color Flicker \& Blinking}
\begin{itemize}[leftmargin=*]
\item Concepts : cyclical color flicker, hinge constraints, multi-frame puzzle, looping animation
\item Description : The input consists of multiple frames showing a hand-drawn scene: a large circular shape with concentric scribbles/colors in its center,  topped by a smaller hinged circle, with various scribbles at the right (ladder-like), left (darker scribble), and near the hinge (green).  Over time, the concentric scribbles in the center circle flicker through different color states while all objects remain in the same relative positions.
\item Full web view is available at  \url{https://gifarc.vercel.app/task/385}.

\begin{figure}[h!]
\centering
    \includegraphics[width=\columnwidth]{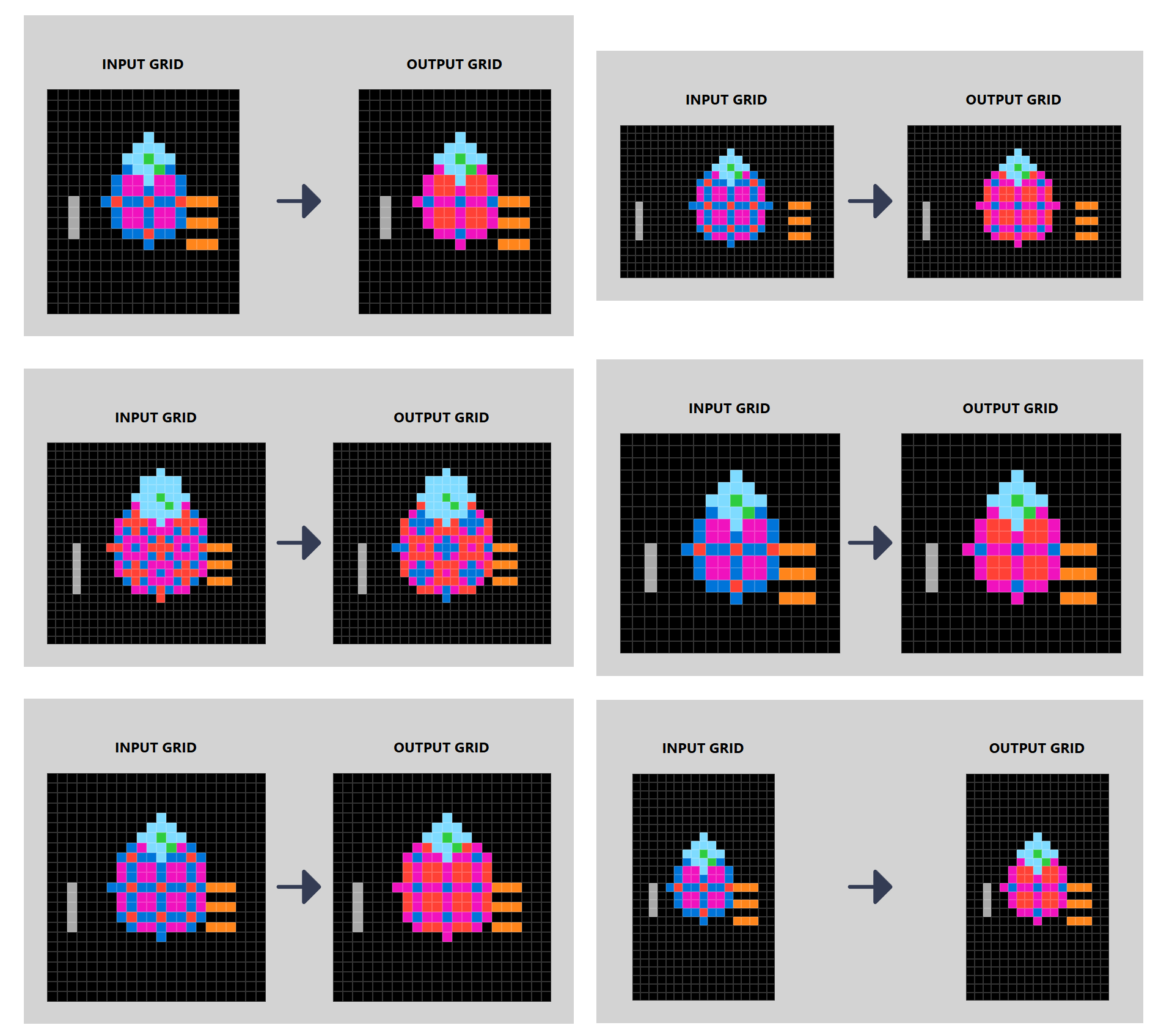}
    \caption{GIFARC-generated task 385.}
    \label{fig:vis_type_7}
\end{figure}

\end{itemize}

\newpage
\subsection{Example Task of Type 8 - Gradient \& Layered Color Changes}
\begin{itemize}[leftmargin=*]
\item Concepts : radial color cycling, symmetry, continuous transformations, layering
\item Description : In the input, you will see a single circular shape on a purely black background. Inside the circle  is a radial, symmetrical pattern composed of multiple colors. Each concentric ring of the circle  shifts colors in a periodic cycle, from an inner ring (earliest phase) to an outer ring (later phase), repeatedly. The circle retains its size and position, and the background remains black.
\item Full web view is available at  \url{https://gifarc.vercel.app/task/117}.

\begin{figure}[h!]
\centering
    \includegraphics[width=\columnwidth]{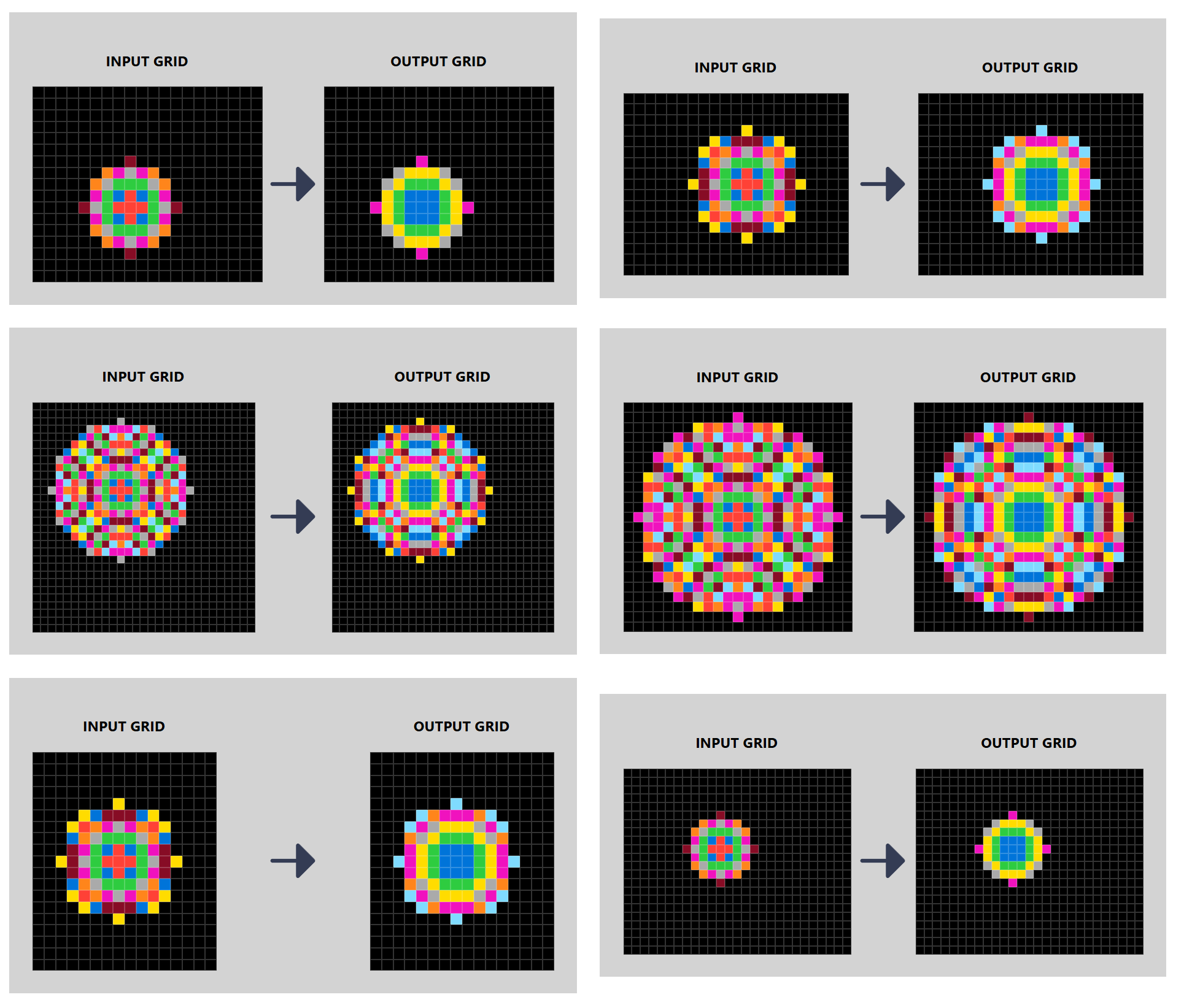}
    \caption{GIFARC-generated task 117.}
    \label{fig:vis_type_8}
\end{figure}

\end{itemize}

\newpage
\subsection{Example Task of Type 9 - Glitch \& Breaking Patterns}
\begin{itemize}[leftmargin=*]
\item Concepts : glitch patterns, color flickering, repetitive animation, partial stability
\item Description : In the input you will see multiple horizontal bars stacked vertically over a black background.  Each bar is composed of green, purple, and yellow pixels arranged in rectangular stripes.  Some small clusters of yellow pixels flicker like digital noise, intensifying toward the center  and right side of each bar. The bars remain in the same positions throughout the sequence.
\item Full web view is available at  \url{https://gifarc.vercel.app/task/410}.

\begin{figure}[h!]
\centering
    \includegraphics[width=\columnwidth]{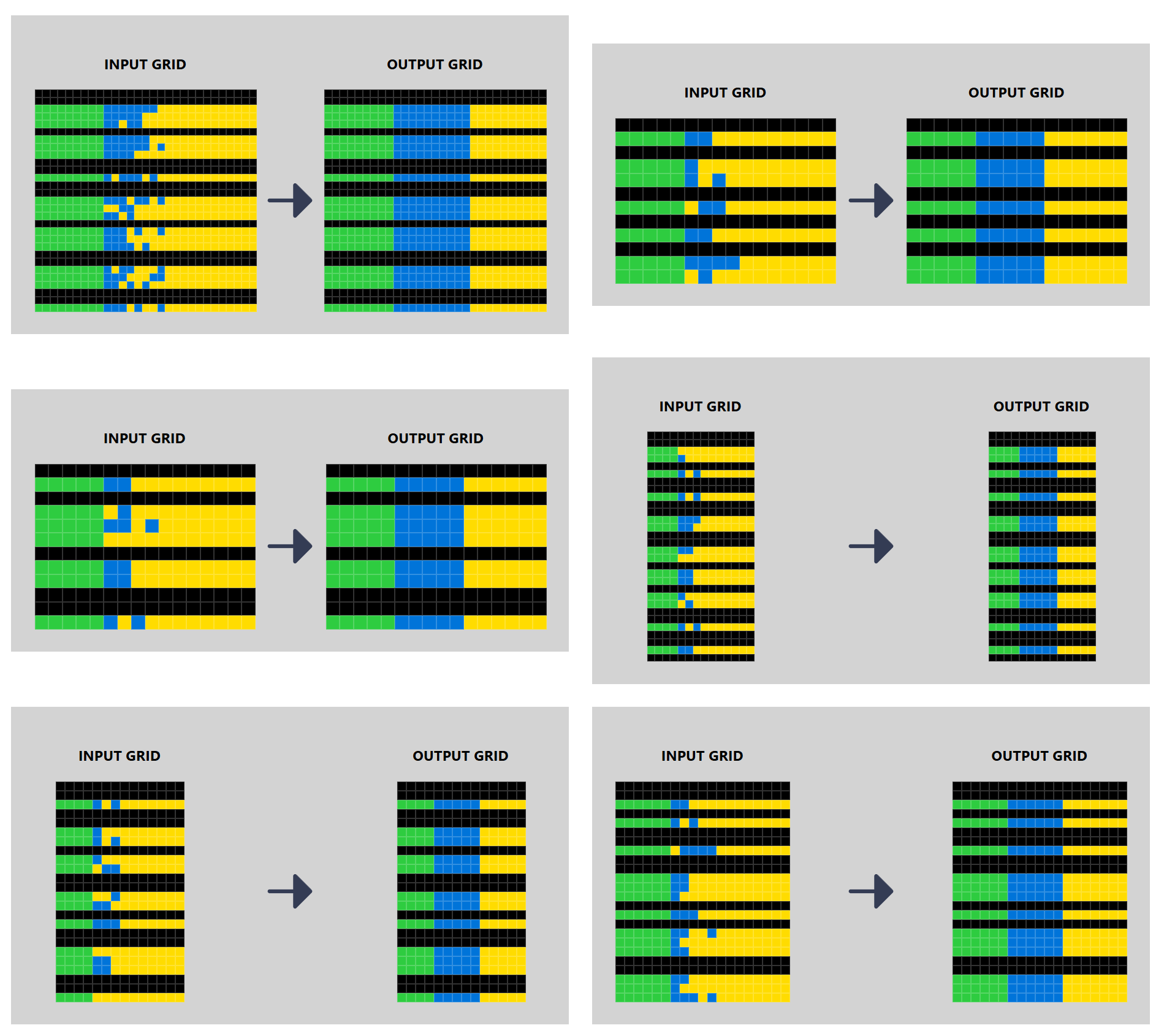}
    \caption{GIFARC-generated task 410.}
    \label{fig:vis_type_9}
\end{figure}

\end{itemize}

\newpage
\subsection{Example Task of Type 10 - Wave \& Diagonal Flow}
\begin{itemize}[leftmargin=*]
\item Concepts : cyclical pulsing, polygonal wave propagation, time-lapse transformation
\item Description : In the input, you will receive a sequence of frames representing a dark polygonal surface with green glowing points at each polygon vertex. Each frame gradually shifts the brightness of the polygons in wave-like motions, causing shimmering highlights that travel along the connected polygon faces. The initial frame and the final frame share the same luminous intensity, forming a seamless loop of pulsing and returning.  Your task:   1. Identify the green points at the vertices, which remain spatially fixed throughout the sequence.   2. Track the wave-like shimmering across the polygon faces.   3. Ensure that the final frame's brightness pattern matches the initial frame's, capturing the cyclical nature of the pulse.   4. Output the entire transformation as a collection of frames (or a metadata structure) that visually loops back to the start.  The key principles are:   - The polygon mesh arrangement does not change; only the luminous intensity on the polygon faces and green vertices fluctuates.   - Shimmering waves spread across the surface repeatedly, then recede, creating a looping timeline.   - The result is a time-lapse style output in which the final state seamlessly resets to the first state, preserving the thought of     continuous pulsing.
\item Full web view is available at  \url{https://gifarc.vercel.app/task/202}.

\begin{figure}[h!]
\centering
    \includegraphics[width=\columnwidth]{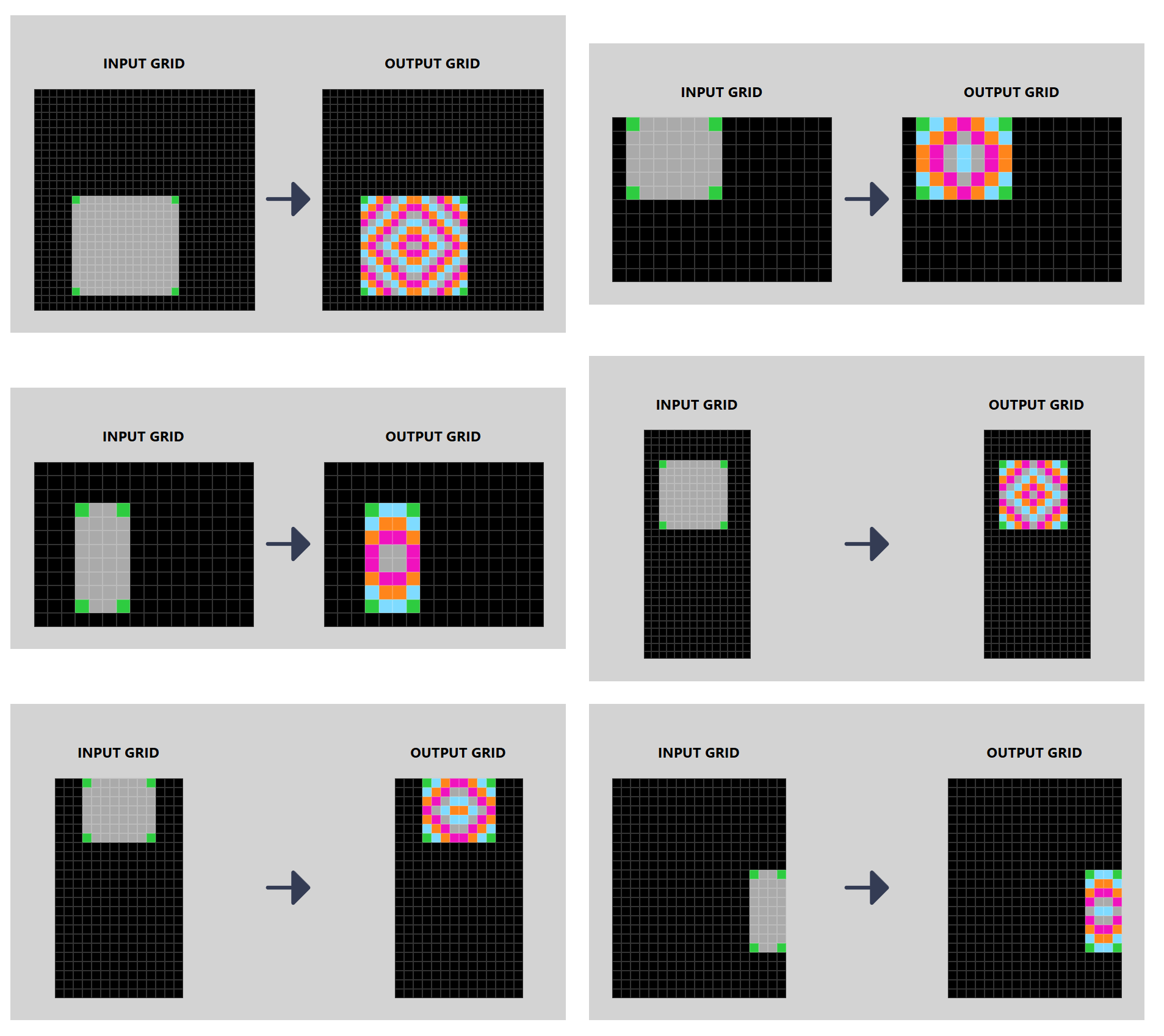}
    \caption{GIFARC-generated task 202.}
    \label{fig:vis_type_10}
\end{figure}

\end{itemize}

\newpage
\subsection{Example Task of Type 11 - Gravity \& Liquid Flow
}
\begin{itemize}[leftmargin=*]
\item Concepts : time-lapse, incremental changes, layering, chart visualization
\item Description : In the input you will see multiple "frames" of a chart with vertical bars and a pink line, each frame on a separate grid.  The black background, rectangular grid, and bounding box remain constant across these frames.  In each subsequent frame, the vertical bars shift in height and flicker in intensity,  while the pink line oscillates up and down with a net upward trend.
\item Full web view is available at  \url{https://gifarc.vercel.app/task/1413}.

\begin{figure}[h!]
\centering
    \includegraphics[width=\columnwidth]{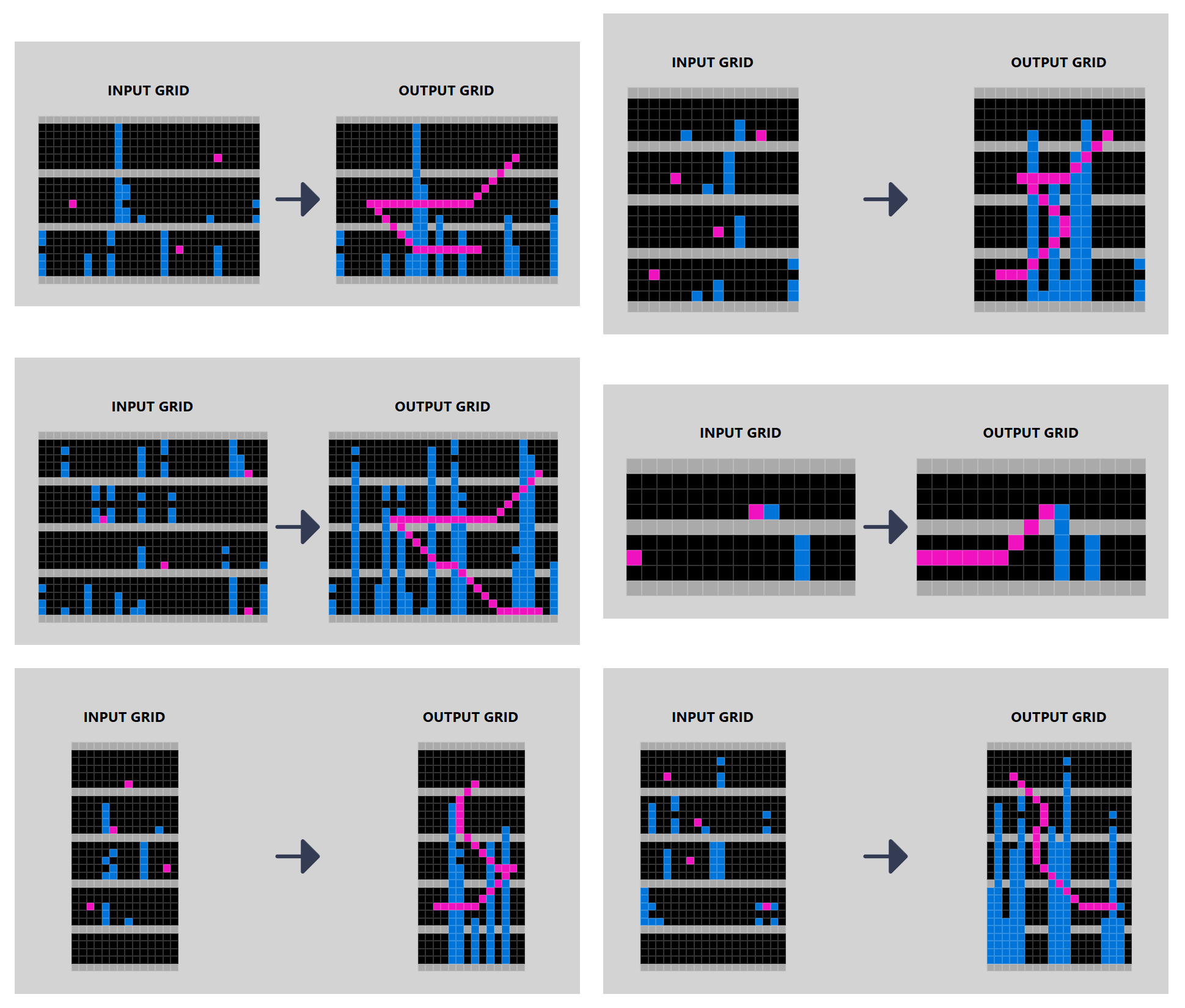}
    \caption{GIFARC-generated task 1413.}
    \label{fig:vis_type_11}
\end{figure}

\end{itemize}

\newpage
\subsection{Example Task of Type 12 - Slow Environmental Change
}
\begin{itemize}[leftmargin=*]
\item Concepts : time-lapse, incremental changes, layering, chart visualization
\item Description : In the input you will see multiple "frames" of a chart with vertical bars and a pink line, each frame on a separate grid.  The black background, rectangular grid, and bounding box remain constant across these frames.  In each subsequent frame, the vertical bars shift in height and flicker in intensity,  while the pink line oscillates up and down with a net upward trend.
\item Full web view is available at  \url{https://gifarc.vercel.app/task/2500}.

\begin{figure}[h!]
\centering
    \includegraphics[width=\columnwidth]{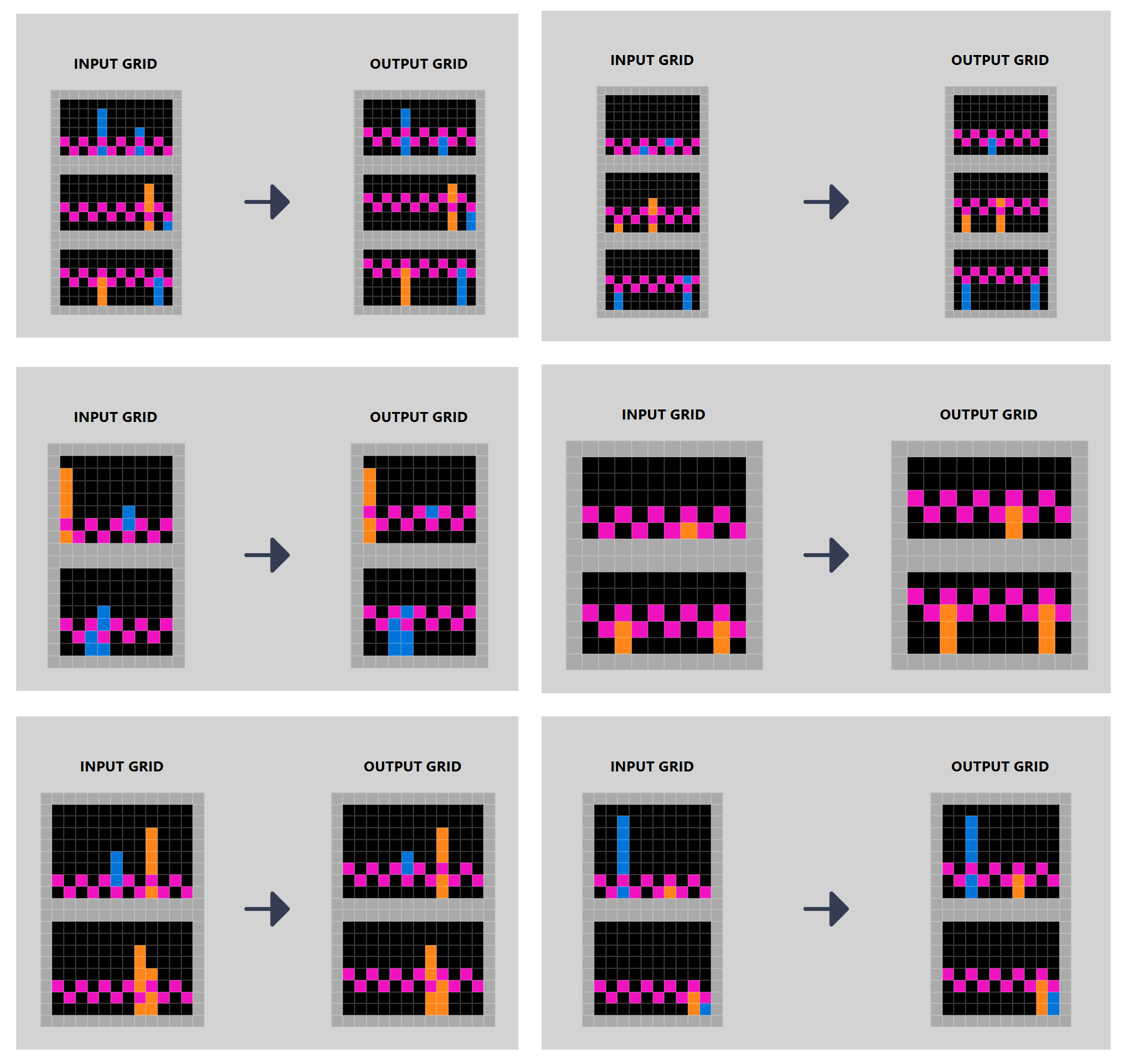}
    \caption{GIFARC-generated task 2500.}
    \label{fig:vis_type_12}
\end{figure}

\end{itemize}

\newpage
\subsection{Example Task of Type 13 - Scaling Burst \& Shape Morphing}
\begin{itemize}[leftmargin=*]
\item Concepts : repetitive scaling, cyclical emergence, timed transformations
\item Description : In the input you will see a pink background with one or more black rings near the center.  Each ring should continuously expand outward from the center until it goes off the grid, at which point it disappears.  Meanwhile, whenever an existing ring crosses halfway toward the border, a new smaller ring is spawned in the center.  This process loops indefinitely—old rings vanish at the boundary, and new rings keep forming in the center.  The output should capture this entire cycle, from the initial state to the eventual large rings that disappear, and newly formed rings that repeat the pattern.  In a static puzzle context, illustrate at least one full cycle of rings moving outward and disappearing, while new rings appear at the center.
\item Full web view is available at  \url{https://gifarc.vercel.app/task/651}.

\begin{figure}[h!]
\centering
    \includegraphics[width=\columnwidth]{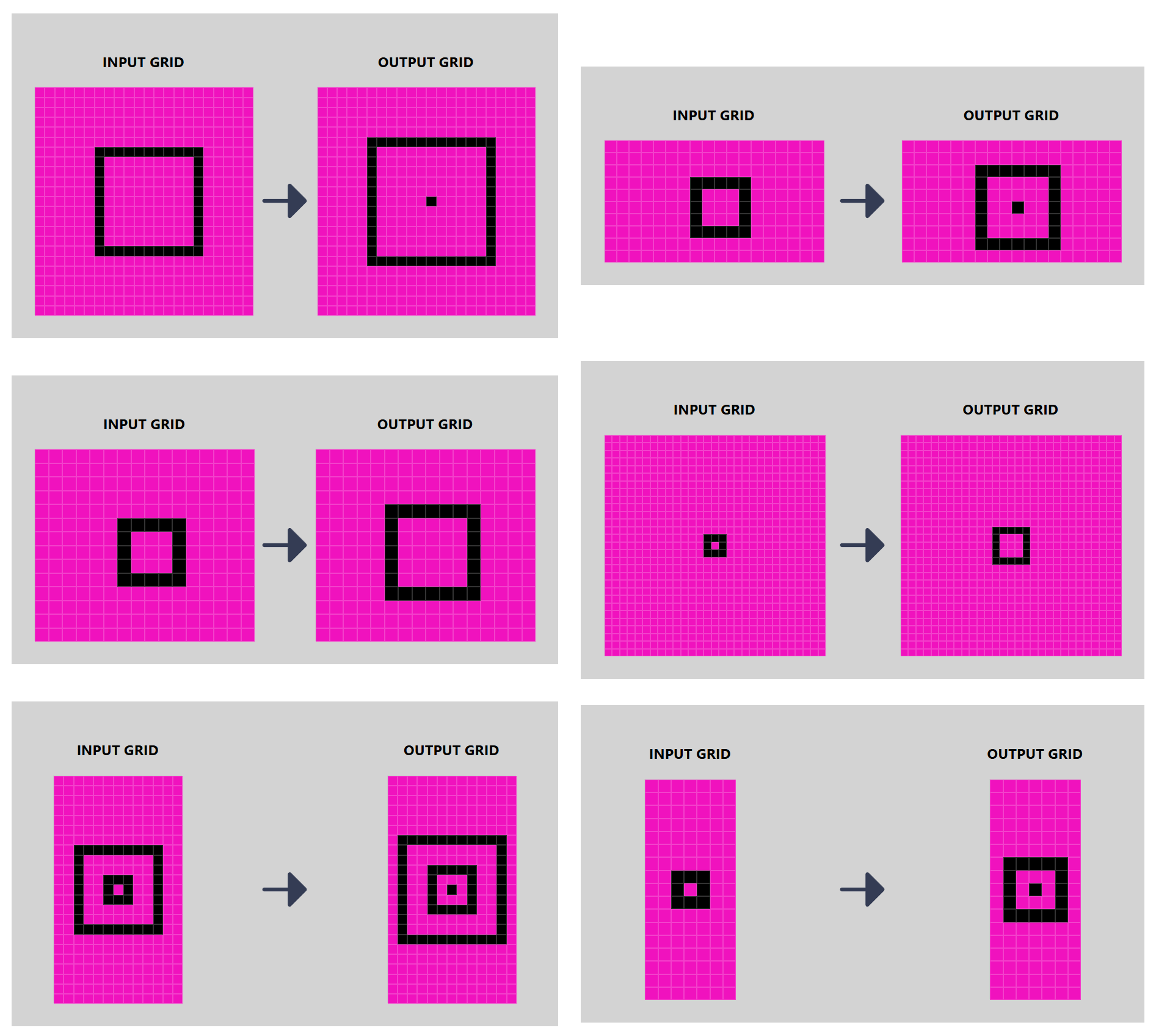}
    \caption{GIFARC-generated task 651.}
    \label{fig:vis_type_13}
\end{figure}

\end{itemize}

\newpage
\subsection{Example Task of Type 14 - Attach/Detach Clusters}
\begin{itemize}[leftmargin=*]
\item Concepts : multi-frame transformation, attachment/detachment, motion, object grouping
\item Description : The input represents a time-sequence (e.g., multiple "frames") showing a suited host seated at a desk with several static background objects (gift bag, city backdrop, building spire). The host reaches for a black mug, and we see that the tie is inadvertently attached to the mug (one frame shows the tie moving upward with the mug). In subsequent frames, the host detaches the tie from the mug and returns both to their normal positions. The core principle to illustrate is that if two objects are "attached"  they move together until they become detached. Your output is the final frame where the tie is no  longer stuck to the mug. The puzzle solution must show how the tie becomes free again, returning  to its intended resting position, while the mug is back on the desk.
\item Full web view is available at  \url{https://gifarc.vercel.app/task/7128}.

\begin{figure}[h!]
\centering
    \includegraphics[width=\columnwidth]{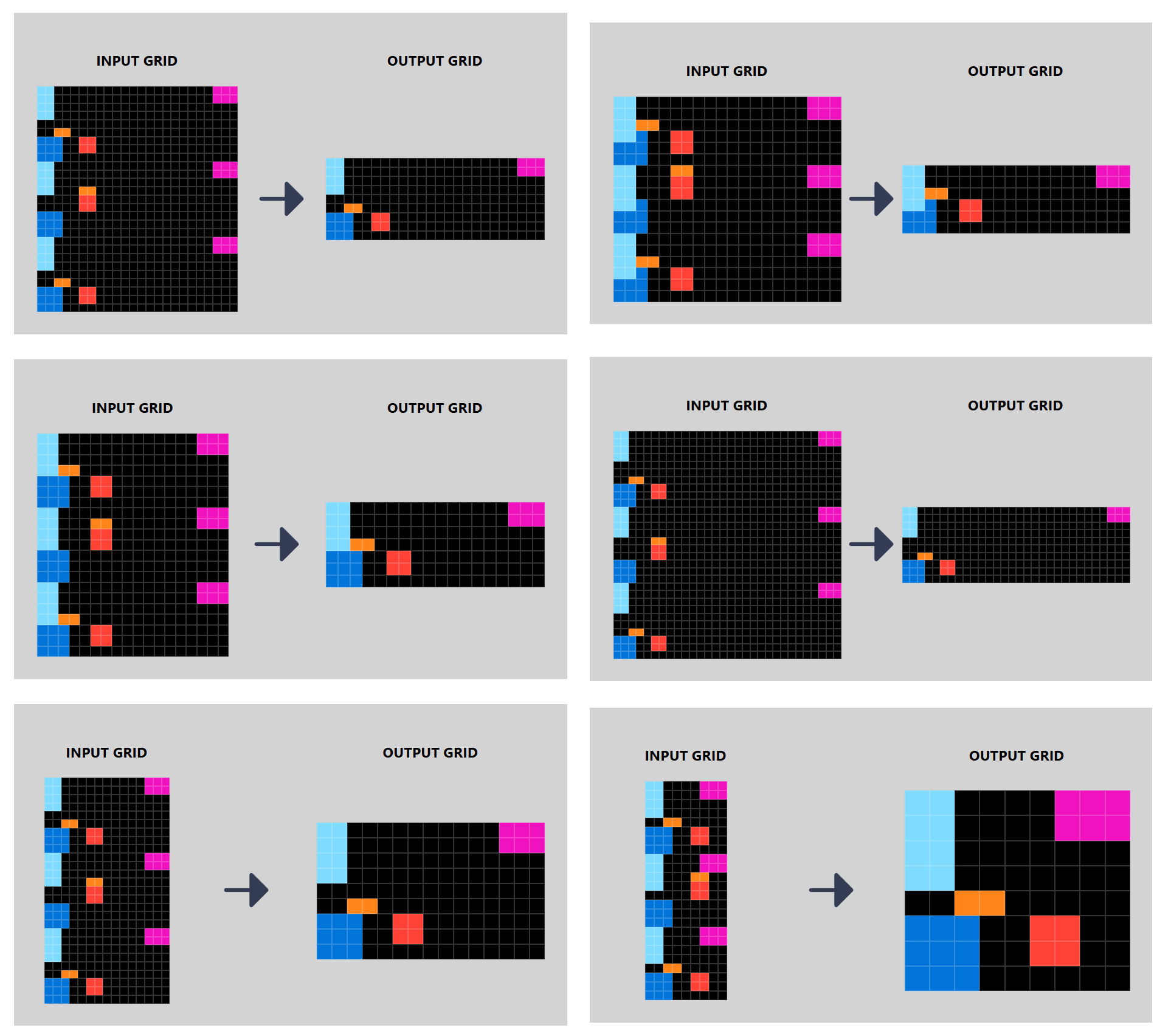}
    \caption{GIFARC-generated task 7128.}
    \label{fig:vis_type_14}
\end{figure}

\end{itemize}

\newpage
\subsection{Example Task of Type 15 - Layer Separation \& Merging}
\begin{itemize}[leftmargin=*]
\item Concepts : color gradient, layered squares, temporal progression, partial transparency
\item Description : In the input you will see a multi-frame sequence of an overlapping grid of squares starting  in dark blues at the top, transitioning gradually to lighter, more yellow tones at the bottom.  The squares remain in a consistent arrangement, but their colors shift slightly from frame  to frame. Overlaps create partial transparency effects that slightly modify the underlying  tones. To produce the output, you must show the final frame in which the color gradient  transitions all the way to lighter shades near the bottom. Capture the entire transformation  by preserving the arrangement of squares, ensuring that no abrupt color changes occur  between consecutive frames, and maintaining a seamless progression of hues through the final  arrangement.
\item Full web view is available at  \url{https://gifarc.vercel.app/task/214}.

\begin{figure}[h!]
\centering
    \includegraphics[width=\columnwidth]{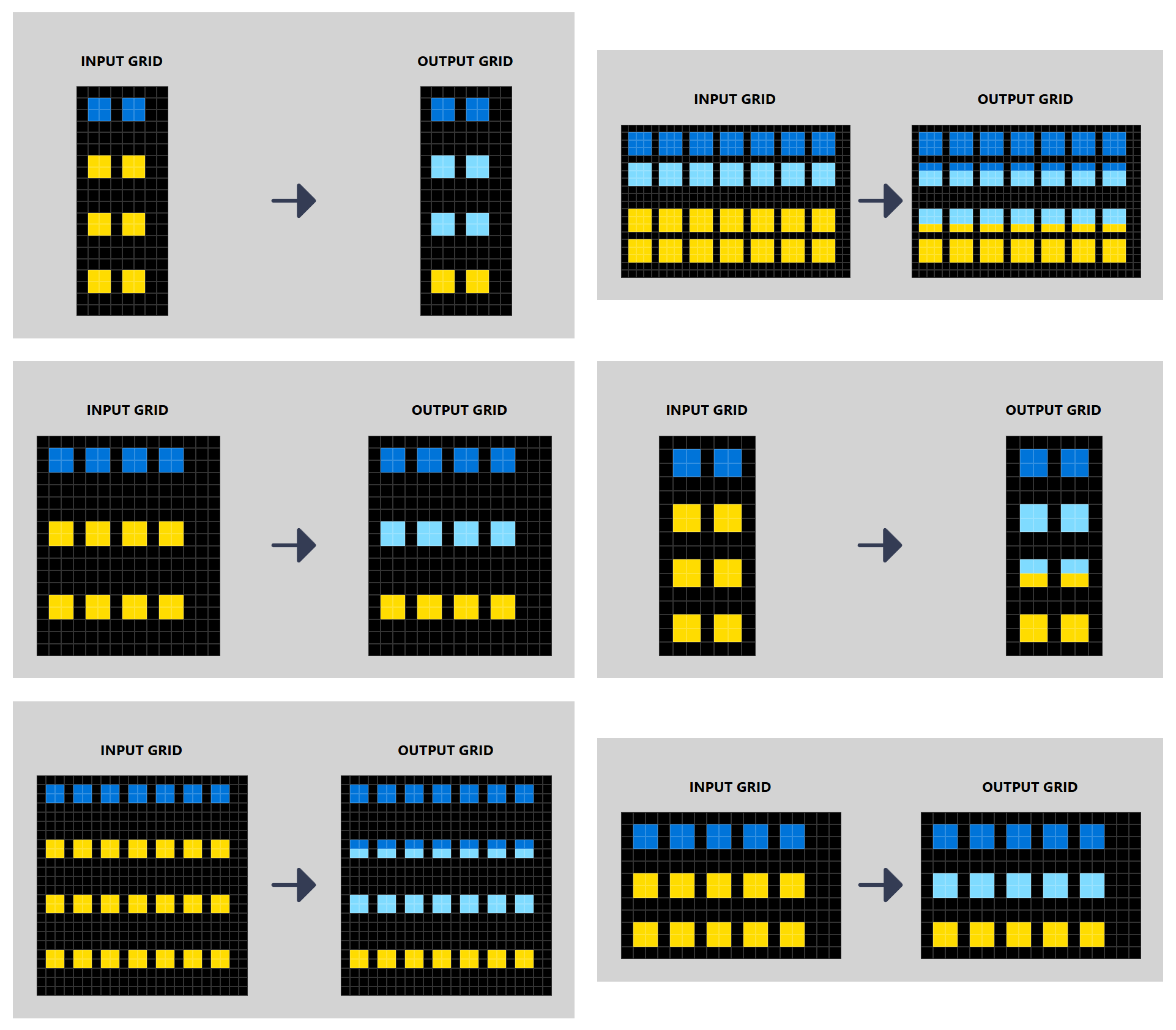}
    \caption{GIFARC-generated task 214.}
    \label{fig:vis_type_15}
\end{figure}

\end{itemize}

\newpage
\subsection{Example Task of Type 16 - Text \& Punctuation Transformation}
\begin{itemize}[leftmargin=*]
\item Concepts : frame-based animation, looping, minimal movement, static overlay
\item Description : In this puzzle, the input consists of multiple frames showing a black background with a rectangular banner. On the left side of the banner is the static text "I'm waiting...", and on the right side is a small clock icon that subtly changes appearance from frame to frame. To produce the output: 1. Keep the banner and text exactly the same in each frame, preserving their positions. 2. Continuously cycle the clock frames in a loop so that it appears to be animating. 3. The clock must remain contained within the banner boundary throughout all frames. 4. The final output (or sequence of frames) should repeat the frames of the clock's movement in an endless loop, preserving the overall layout within each frame.
\item Full web view is available at  \url{https://gifarc.vercel.app/task/2222}.

\begin{figure}[h!]
\centering
    \includegraphics[width=\columnwidth]{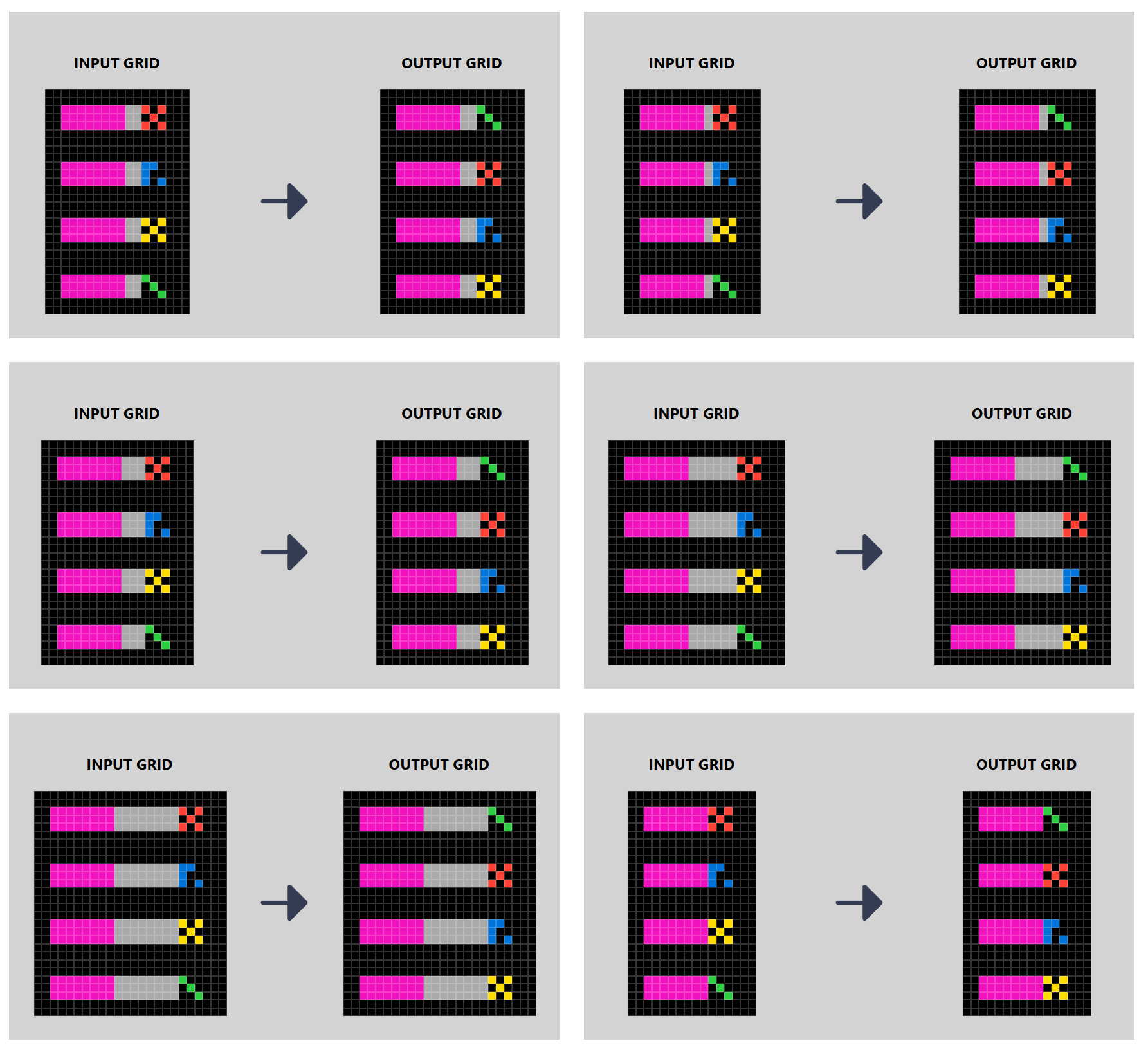}
    \caption{GIFARC-generated task 2222.}
    \label{fig:vis_type_16}
\end{figure}

\end{itemize}

\newpage
\subsection{Example Task of Type 17 - Minimal Motion Overlay}
\begin{itemize}[leftmargin=*]
\item Concepts : multi-frame static verification, scene consistency, no-change detection
\item Description : In the input, you will see several frames that together depict a stylized scene:   - A top sphere positioned at the upper center   - A curved band placed directly below the sphere   - Two large symmetrical side shapes in yellow   - A vertical pink line in the center   - A signature text in the bottom right corner These elements are repeated exactly across all frames with no movement or color transitions. To make the output, you must check if all frames are truly identical. If they are, return a single-frame grid replicating the scene exactly.  If there is any discrepancy (in shape, color, or position) among the frames, then the output should be a black canvas of the same size, indicating the scenes are not perfectly static.
\item Full web view is available at  \url{https://gifarc.vercel.app/task/7262}.

\begin{figure}[h!]
\centering
    \includegraphics[width=\columnwidth]{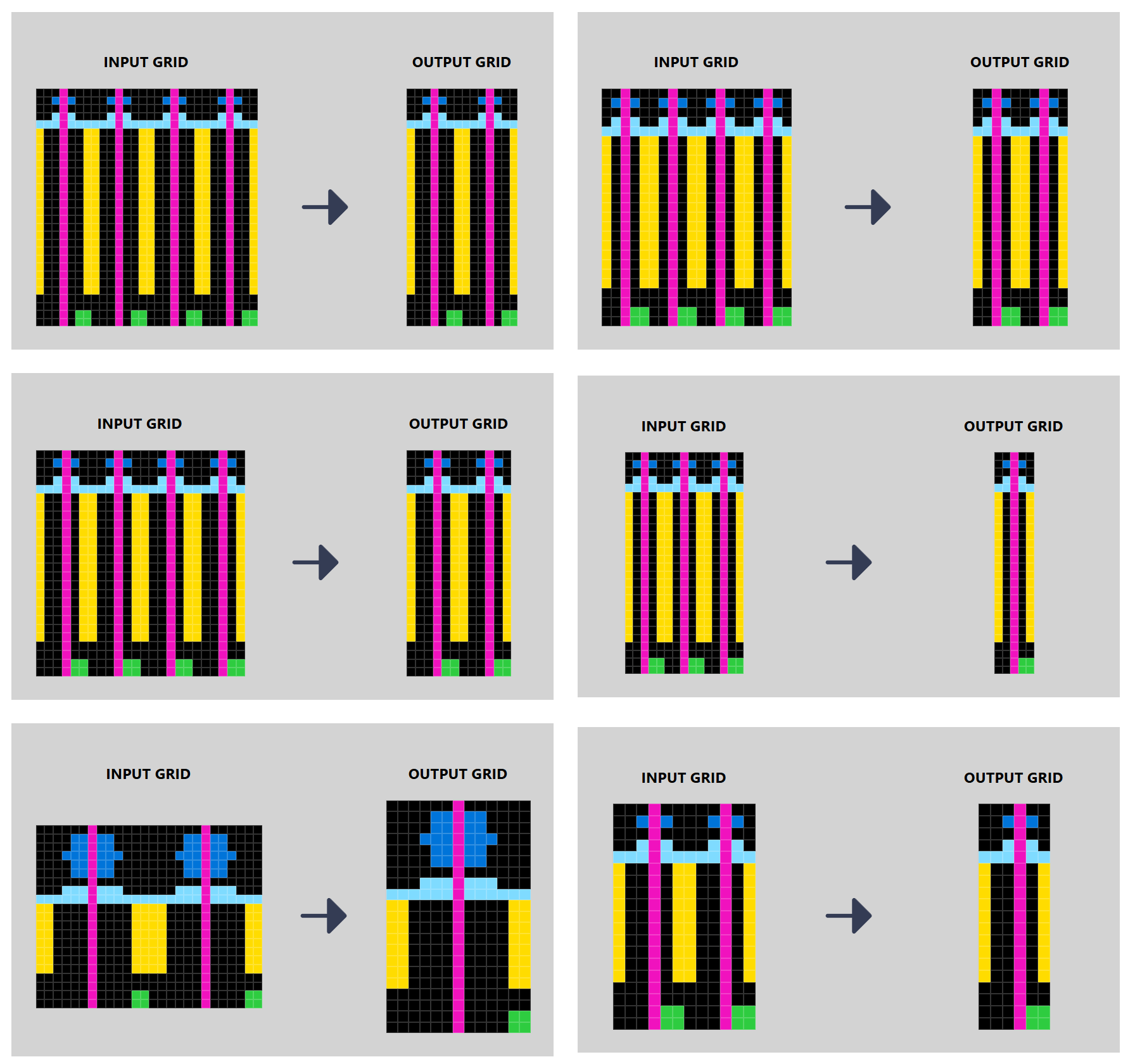}
    \caption{GIFARC-generated task 7262.}
    \label{fig:vis_type_17}
\end{figure}

\end{itemize}

\newpage
\subsection{Example Task of Type 18 - Static Verification \& No Change}
\begin{itemize}[leftmargin=*]
\item Concepts : multi-frame static verification, scene consistency, no-change detection
\item Description : In the input, you will see several frames that together depict a stylized scene:   - A top sphere positioned at the upper center   - A curved band placed directly below the sphere   - Two large symmetrical side shapes in yellow   - A vertical pink line in the center   - A signature text in the bottom right corner These elements are repeated exactly across all frames with no movement or color transitions. To make the output, you must check if all frames are truly identical. If they are, return a single-frame grid replicating the scene exactly.  If there is any discrepancy (in shape, color, or position) among the frames, then the output should be a black canvas of the same size, indicating the scenes are not perfectly static.
\item Full web view is available at  \url{https://gifarc.vercel.app/task/413}.

\begin{figure}[h!]
\centering
    \includegraphics[width=\columnwidth]{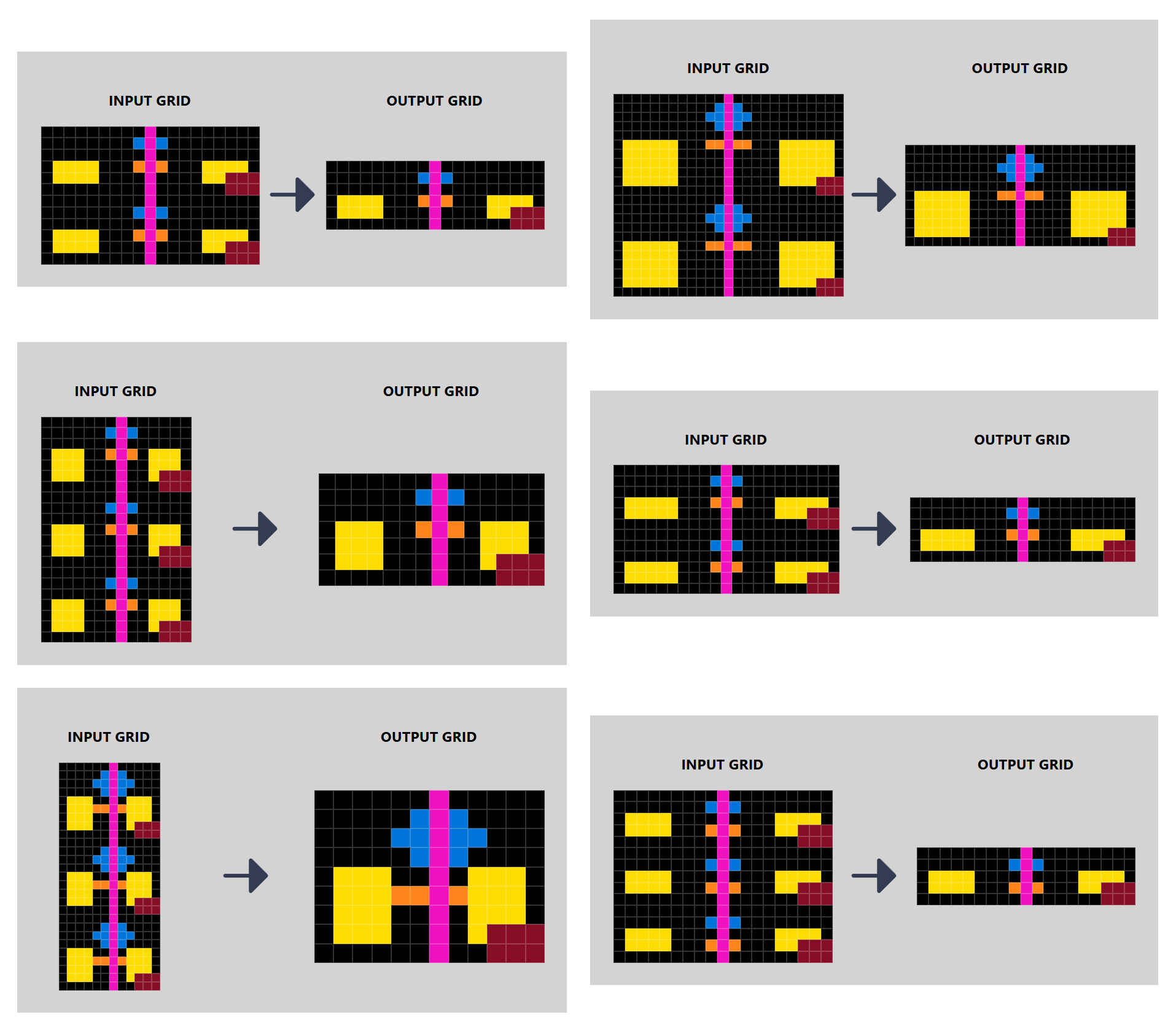}
    \caption{GIFARC-generated task 413.}
    \label{fig:vis_type_18}
\end{figure}

\end{itemize}

\newpage
\subsection{Example Task of Type 19 - Fractal Expansion \& Self-Similar Repeats}
\begin{itemize}[leftmargin=*]
\item Concepts : fractal expansions, symmetrical pulsations, radial transformations, iterative growth
\item Description : In the input, you will see a sequence of grids (frames) on a black background.  Each grid depicts:  - A large circular region in the center, with multiple concentric rings and radial lines diverging outward.  - A stacked-curve fractal anchored at the bottom-left corner.  - A set of spire fractals anchored at the bottom-right corner.  As you move from one frame to the next in the input, these elements undergo iterative transformations:  - The radial lines repeatedly shift in thickness and visual intensity, while preserving rotational symmetry.  - The concentric rings in the circular region pulsate by alternately expanding and contracting.  - The stacked-curve fractal on the bottom-left changes its curve density, fractally adding or removing segments.  - The right-side spire fractals expand and contract in repeated vertical segments.  Your task is to replicate and apply these transformations for the entire sequence, and produce the final frame of the animation as the output:  1) Ensure the anchoring of the fractals at the bottom edge remains the same.  2) Preserve the radial symmetry of the central circular region and its common center point.  3) For each pulsation step in the input, magnify or contract the rings, lines, and fractals accordingly.  4) Continue until the final pulsation step is reached. That final state is your output grid.  The essential principle is that symmetry-based fractal transformations and repeated cyclical expansions/contractions produce the overall pulsating effect.  By reflecting each iterative change step by step, you reconstruct the final pulsating pattern visible at the end of the sequence.
\item Full web view is available at  \url{https://gifarc.vercel.app/task/155}.

\begin{figure}[h!]
\centering
    \includegraphics[width=\columnwidth]{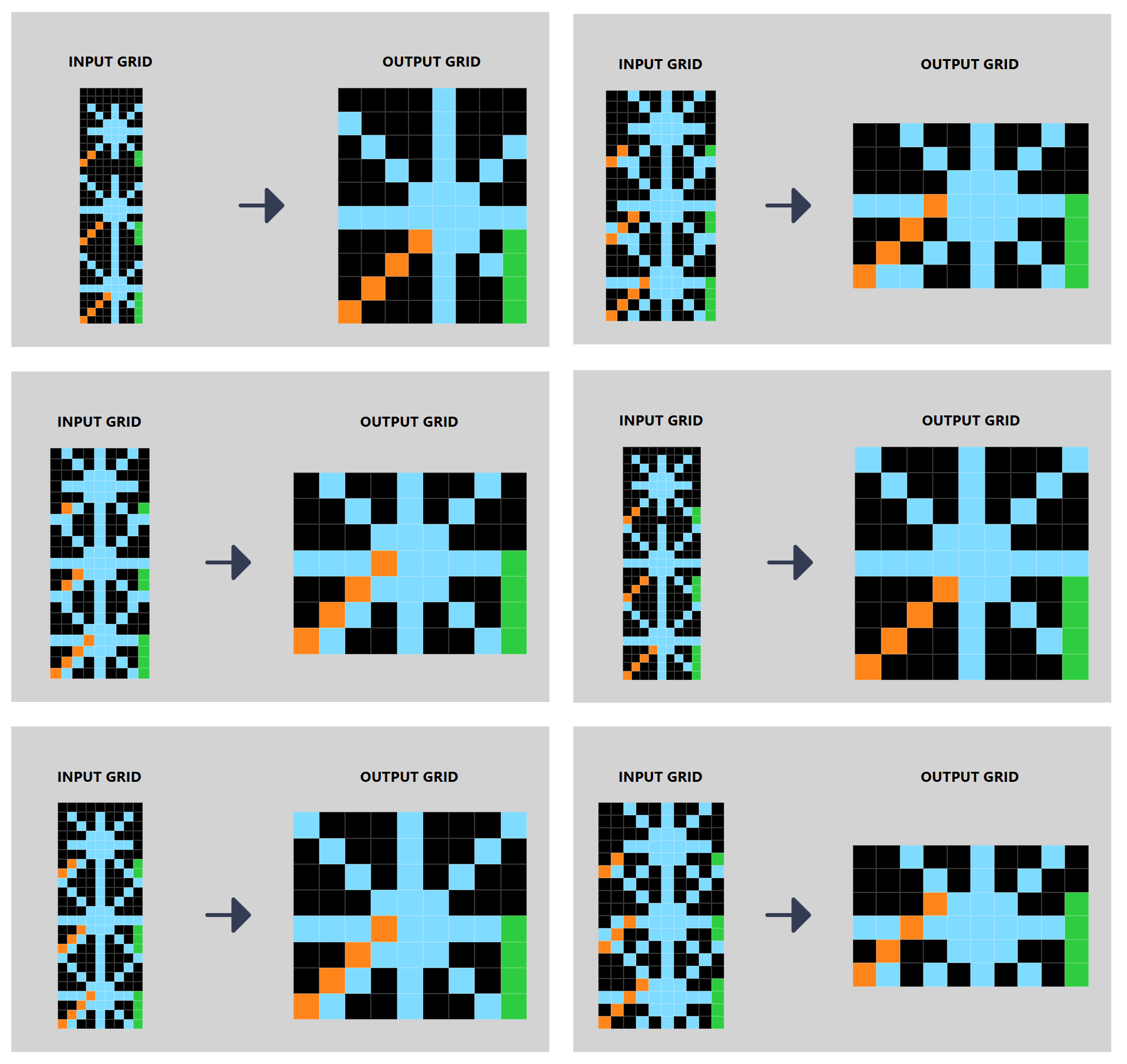}
    \caption{GIFARC-generated task 155.}
    \label{fig:vis_type_19}
\end{figure}

\end{itemize}

\newpage
\subsection{Example Task of Type 20 - Sequential Pattern Growth \& Transition}
\begin{itemize}[leftmargin=*]
\item Concepts : fractal expansions, symmetrical pulsations, radial transformations, iterative growth
\item Description : In the input, you will see a sequence of grids (frames) on a black background.  Each grid depicts:  - A large circular region in the center, with multiple concentric rings and radial lines diverging outward.  - A stacked-curve fractal anchored at the bottom-left corner.  - A set of spire fractals anchored at the bottom-right corner.  As you move from one frame to the next in the input, these elements undergo iterative transformations:  - The radial lines repeatedly shift in thickness and visual intensity, while preserving rotational symmetry.  - The concentric rings in the circular region pulsate by alternately expanding and contracting.  - The stacked-curve fractal on the bottom-left changes its curve density, fractally adding or removing segments.  - The right-side spire fractals expand and contract in repeated vertical segments.  Your task is to replicate and apply these transformations for the entire sequence, and produce the final frame of the animation as the output:  1) Ensure the anchoring of the fractals at the bottom edge remains the same.  2) Preserve the radial symmetry of the central circular region and its common center point.  3) For each pulsation step in the input, magnify or contract the rings, lines, and fractals accordingly.  4) Continue until the final pulsation step is reached. That final state is your output grid.  The essential principle is that symmetry-based fractal transformations and repeated cyclical expansions/contractions produce the overall pulsating effect.  By reflecting each iterative change step by step, you reconstruct the final pulsating pattern visible at the end of the sequence.
\item Full web view is available at  \url{https://gifarc.vercel.app/task/2061}.

\begin{figure}[h!]
\centering
    \includegraphics[width=\columnwidth]{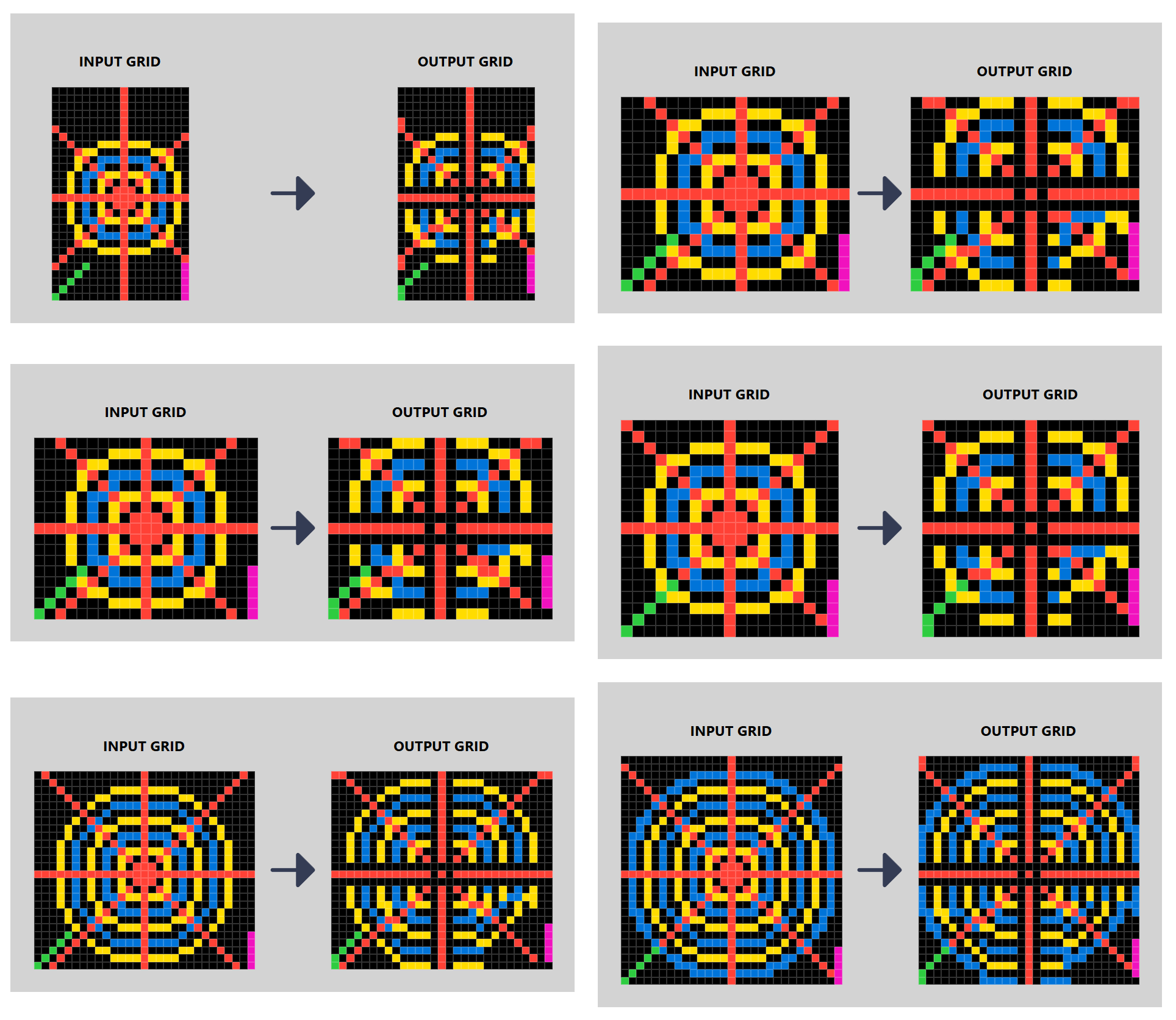}
    \caption{GIFARC-generated task 2061.}
    \label{fig:vis_type_20}
\end{figure}

\end{itemize}




\end{document}